%% file: sn-article.tex
\documentclass[pdflatex,sn-mathphys-num]{sn-jnl}

\usepackage{graphicx}%
\usepackage{multirow}%
\usepackage{amsmath,amssymb,amsfonts}%
\usepackage{amsthm}%
\usepackage{mathrsfs}%
\usepackage[title]{appendix}%
\usepackage{xcolor}%
\usepackage{textcomp}%
\usepackage{manyfoot}%
\usepackage{booktabs}%
\usepackage{algorithm}%
\usepackage{algorithmicx}%
\usepackage{algpseudocode}%
\usepackage{listings}%

\usepackage{siunitx}
\usepackage{adjustbox}
\usepackage[table]{xcolor}
\usepackage[normalem]{ulem}
\usepackage{makecell}
\usepackage{float}
\usepackage{bibunits}
\usepackage{lineno}
\defaultbibliographystyle{sn-mathphys-num}

\usepackage{soul}
\newcommand{\addtext}[1]{\textcolor{black}{#1}}

\usepackage{docmute}

\theoremstyle{thmstyleone}%
\theoremstyle{thmstyletwo}%

\theoremstyle{thmstylethree}%

\renewcommand{\orcidlogo}{%
    \raisebox{0.15em}{%
        \includegraphics[height=0.75em]{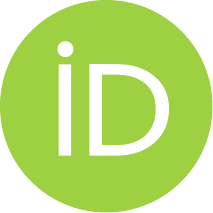}%
    }%
}

\begin{document}

\title[Article Title]{\centering
Non-frontal face recognition using GANs\\and memristor-based classifiers
}

\author*{\fnm{Semih} \sur{Vazgecen} \orcid{https://orcid.org/0009-0003-8868-3423}\textsuperscript{*}}\email{S.A.Vazgecen@sms.ed.ac.uk}

\author{\fnm{Cristian} \sur{Sestito} \orcid{https://orcid.org/0000-0002-7731-0002}}

\author{\fnm{Spyros} \sur{Stathopoulos} \orcid{https://orcid.org/0000-0002-0833-6209}}

\author{\fnm{Themis} \sur{Prodromakis} \orcid{https://orcid.org/0000-0002-6267-6909}}

\affil{\orgdiv{Centre for Electronics Frontiers, Institute for Integrated Micro and Nano Systems},\\ \orgname{School of Engineering, The University of Edinburgh}, \orgaddress{\country{UK}}}


\abstract{Face recognition systems have advanced significantly through deep learning techniques, delivering high performance and robustness in complex scenarios. However, these approaches incur substantial computational overhead, limiting their in situ applicability in resource-constrained platforms such as drones, where they can address challenges including non-frontal facial imagery. Memristor-based neuromorphic systems have emerged as a compelling approach for edge AI applications, combining biologically inspired processing with efficient and scalable computation. In this work, we propose a facial recognition framework that addresses non-frontal pose variations by integrating lightweight generative adversarial network (GAN)-based pose frontalisation with memristor-based neuromorphic recognition. The experimental results on two datasets demonstrate the effectiveness of combining adversarial learning with memristive technology, achieving up to 96\% identification accuracy. The proposed approach alleviates the computational bottlenecks of conventional AI and offers a scalable, efficient solution for face recognition in dynamic real-world environments.}

\maketitle

\section{Introduction}\label{sec1}

Face recognition has emerged from the longstanding need to identify individuals for various purposes, including fraud detection \cite{Wang_Zhan_Zhan_Xu_Bai_2024}, authentication and access control \cite{RAMESWARI20211251}, surveillance \cite{Srivastava}, forensic investigation \cite{Zeinstra}, and border security \cite{8489113}. Despite its widespread adoption and utility, each implementation of face recognition systems suffers from inherent challenges and limitations in practical deployments. Typically, individuals provide frontal biometric images to the relevant authorities, and these images commonly serve as reference templates for custom application contexts. However, in real-world scenarios, it is not always possible to capture front-facing images of individuals of interest. In some deployments, the acquisition process is relatively straightforward, relying on static ground-based cameras operating under controlled imaging conditions, such as those used in smartphone face-unlock systems \cite{10150/672890} or ePassport gates \cite{SANCHEZDELRIO201649} at airports. Beyond these constrained and stationary identification settings, emerging surveillance applications increasingly require mobile and adaptive recognition systems capable of operating in dynamic, large-scale environments. In such scenarios, facial imagery is frequently acquired under unconstrained and suboptimal conditions, where environmental factors substantially degrade recognition accuracy. Moreover, even under favourable imaging conditions, individuals may deliberately attempt to evade detection through partial facial occlusion \cite{app15179390, cheng2018surveillancefacerecognitionchallenge}. Deep learning approaches offer substantial potential to address these challenges through generative augmentation, robust feature extraction, and domain-adaptive learning strategies \cite{8237529, 7477555, 8550772}. However, enhanced machine cognition incurs substantial computational costs, which poses a major challenge for edge platforms operating under low size, weight, and power (SWaP) constraints, such as drones \cite{9778241}. To alleviate this challenge, various strategies have been proposed \cite{Suganya, Dadheech2025AirNavACC, 7842423, https://doi.org/10.1155/2021/8615367}; yet these approaches remain less suitable for mission-critical applications involving sensitive data, such as law enforcement and defence, and are impractical in environments with limited or unreliable connectivity.

In order to mitigate the growing computational burden, post-CMOS devices are progressively replacing conventional von Neumann architectures. This has driven the development of emerging computing paradigms, including neuromorphic computing, which seek to overcome the limitations above. Physical computing substrates have also been combined with neural-network-based decision-making in other recognition contexts, such as optical correlation systems followed by neural-network-based classification \cite{MOHAPATRA2026109719}. Among these emerging technologies, memristors hold substantial promise for enabling efficient neural computation on edge devices. These two-terminal analogue devices exhibit non-volatile resistive states and low-power operation, making them suitable for in-memory and neuromorphic computing architectures \cite{HuangYi2024, nano15141130, aguirrefernando2024}. Memristive neuromorphic systems, often in conjunction with spiking neural networks (SNNs), have demonstrated promising performance in tasks such as classification \cite{10666111, doi:10.1142/S0218348X23400406} and generation \cite{mi15020217, 8832460, liu2019memristorbasedunsupervisedneuromorphic}.

In this work, we propose a face recognition framework that addresses the non-frontal pose problem and targets resource-constrained edge environments. The framework integrates two complementary strategies. First, a convolutional generative adversarial network (GAN) is employed for face reconstruction, synthesising frontal facial images from non-frontal captures to mitigate self-occlusion caused by the angle between the camera and the subject. Second, memristor-based classifiers\addtext{, empirically calibrated using in-house fabricated devices,} are leveraged for face recognition, processing the synthetic frontal images to efficiently and accurately identify the target. In addition to identifying known individuals, the framework also accommodates an auxiliary mechanism for unknown rejection and online learning, as real-world recognition scenarios are inherently open-set and continuously evolving. Together, these techniques aim to harness the potential of AI for mobile and resource-constrained face recognition applications. Evaluated on two independent benchmark datasets, our framework demonstrates significant improvements in identification accuracy, achieving average recognition rates of 96.3\% on the CMU Multi-PIE dataset \cite{4813399} and 73.4\% on the DroneFace dataset \cite{10.1145/3083187.3083214}.

\section{Results}\label{sec2}

\subsection{Impact of pose variability on face recognition performance}

\begin{figure}
  \begin{center}
  \includegraphics[width=\linewidth]{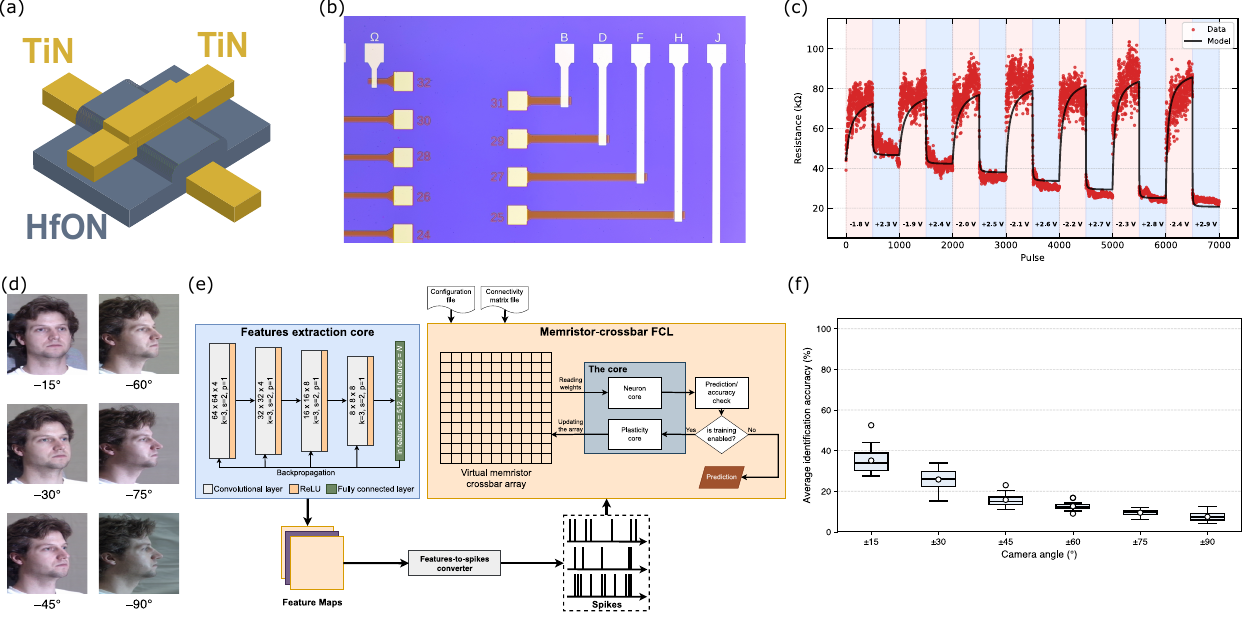}\\
  \caption{\textbf{Non-frontal face classification using SNN}. \textbf{(a)} Schematic representation of a HfON-based memristor device. \textbf{(b)} Optical micrograph of devices. \textbf{(c)} Empirical model fitting. Analytical model (solid black line) fitted to the experimentally measured resistive response of TiN/HfON/TiN device (red dots), obtained from 500-pulse batches of alternating-polarity stimuli. The amplitudes of the applied pulse trains are indicated along the bottom axis of the graph. \textbf{(d)} Non-frontal facial image samples acquired from multiple camera viewpoints in the original CMU Multi-PIE dataset \cite{4813399}. \textbf{(e)} Schematic diagram of the face classification unit (FCU). The FCU consists of three main components: a features extraction core based on a conventional convolutional neural network (CNN) for robust feature representation; a features-to-spike converter that encodes feature maps into spike trains according to the selected encoding scheme; and a memristor crossbar fully connected layer (FCL), which forms the core of the FCU and implements a spiking neural network (SNN) by mapping spike-encoded features onto a memristor crossbar array and updating synaptic weights according to the chosen neuron model and learning rule. \addtext{\textbf{(f)}} Average identification accuracy per camera angle bin of the Multi-PIE dataset for the FCU.}\label{main_figure_1}
  \end{center}
\end{figure}

To investigate the impact of pose variance on face recognition performance, we evaluated the proposed memristor-based face classification unit (FCU) on the CMU Multi-PIE dataset \cite{4813399}. The FCU is a spiking neural network (SNN) implemented on a memristor-based simulation platform that provides algorithm-level modelling for design, testing, and validation in \addtext{a} hardware-representative environment prior to deployment \cite{Huang_2022, 10666111}. We developed an empirical memristor model from experimental measurements to enable robust data-driven representation of memristive behaviour throughout the study. Figure \ref{main_figure_1}(a) presents a representative device schematic, while Figure \ref{main_figure_1}(b) shows an optical micrograph of the fabricated TiN/HfON/TiN devices used for characterisation. Figure \ref{main_figure_1}(c) shows the parameter-fitting results of the empirical memristor model across voltage-amplitude windows. Experimental data points (red dots) are overlaid with model predictions (solid black line), demonstrating close agreement and accurate reproduction of the measured resistance evolution.

In this experiment, we trained the FCU using frontal images of 50 selected subjects from the CMU Multi-PIE dataset. For inference, non-frontal images of the same subjects were grouped into angle bins according to camera orientation. Representative samples from selected angle bins are shown in Figure \ref{main_figure_1}(d). Figure \ref{main_figure_1}(e) illustrates the FCU architecture. Starting from image feature extraction, the resulting feature maps are converted into spike trains and used for memristor-based SNN training, following the chosen neuron models and learning rules. The training configuration for this experiment is summarised in Supplementary Table S1. Supplementary Figure S7 illustrates key aspects of the memristor-based training process, including the evolution of device memristive states over the training session, the pre- and post-training conductance distributions, and the spike activity history. Further details regarding these parameters are provided in the Methods section.

The average identification accuracies for the non-frontal images are shown in Figure \ref{main_figure_1}\addtext{(f)}. The results show a consistent decline in classification accuracy with increasing pose angle. The highest accuracy is achieved at ±15° (35.13\% ± 6.35\%), followed by ±30° (25.80\% ± 5.03\%). A significant drop is observed at ±45° (15.80\% ± 3.07\%), with performance continuing to decrease at ±60° (12.63\% ± 1.83\%), ±75° (9.46\% ± 1.49\%), and reaching 7.44\% ± 2.04\% at ±90°. These results demonstrate the sensitivity of the proposed system to pose variations, with accuracy degrading as the viewing angle deviates from the frontal position. This trend is further supported by the confusion matrices (see Supplementary Figure S10), which reveal a gradual dispersion of predictions away from the diagonal, with increasing off-diagonal entries at larger pose angles.

\begin{figure}
  \begin{center}
  \includegraphics[width=\linewidth]{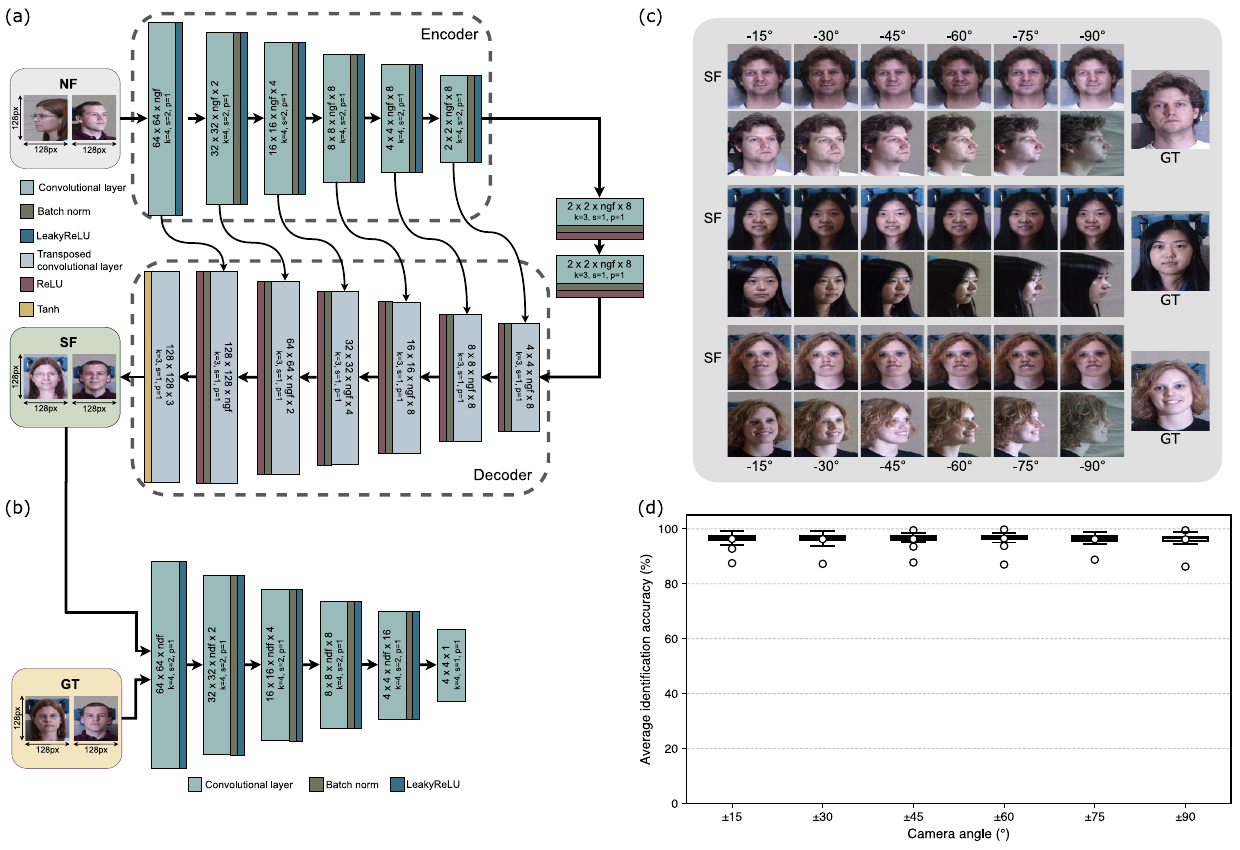}\\
  \caption{\textbf{Face classification using GAN-based reconstructed images.} \textbf{(a)} and \textbf{(b)} illustrate the two main components of the face reconstruction unit (FRU). \textbf{(a)} Generator architecture of the FRU, consisting of an encoder–decoder structure with U-Net–style skip connections that reconstruct non-frontal (NF) input images into their synthetic frontal (SF) versions. \textbf{(b)} Discriminator architecture of the FRU, designed to distinguish between ground-truth (GT) frontal images and SF images synthesised by the Generator. \textbf{(c)} Frontal view synthesis results of the FRU. For each individual, the top row shows the SF images generated from the NF input images in the bottom row. GT frontal images are provided on the right for reference. \textbf{(d)} Average identification accuracy per camera-angle bin on the CMU Multi-PIE dataset \cite{4813399} using the face classification unit (FCU) applied to FRU-reconstructed images.}\label{main_figure_2}
  \end{center}
\end{figure}

\begin{figure}
  \begin{center}
  \includegraphics[width=\linewidth]{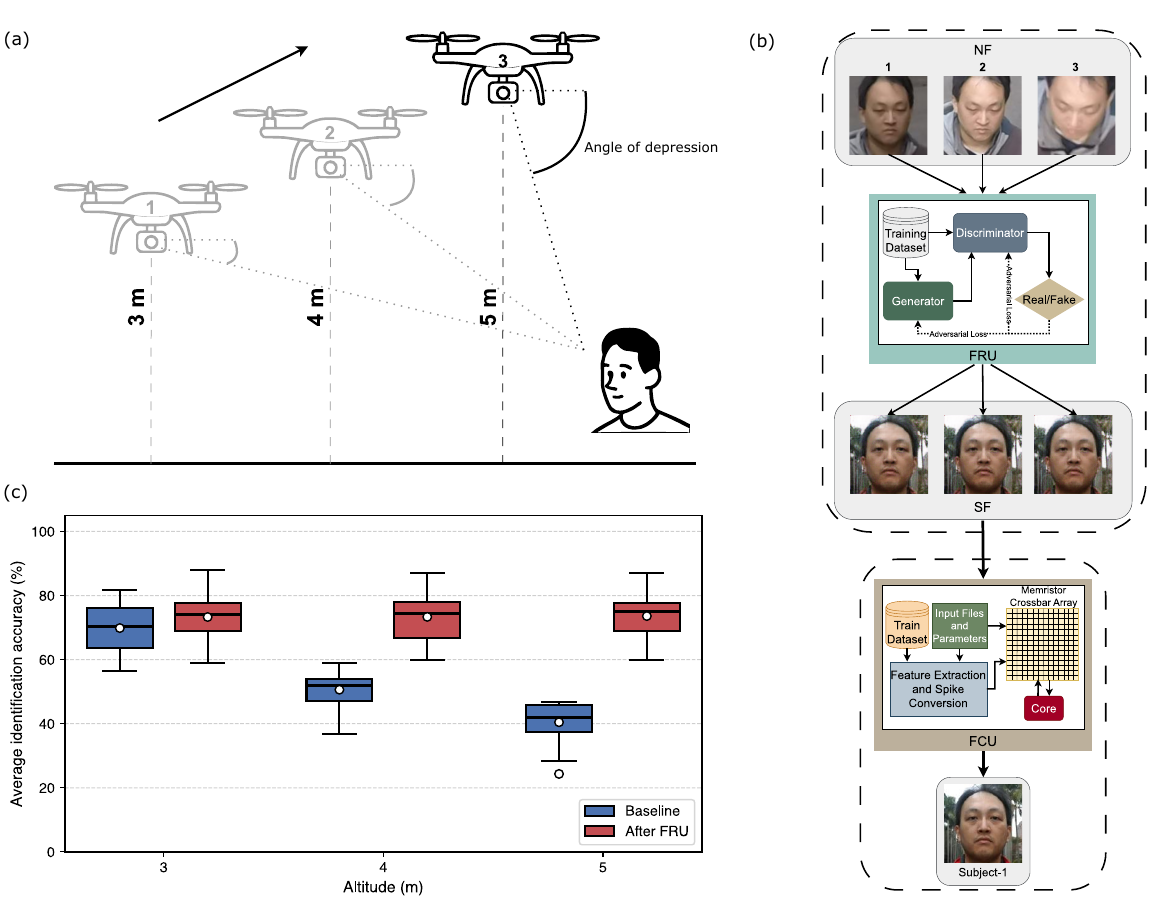}\\
  \caption{\textbf{Face classification of drone-captured \addtext{images} using FRU reconstruction.} \textbf{(a)} Illustration of the creation methodology of the DroneFace dataset  \cite{10.1145/3083187.3083214}. A drone captures sequential images of individuals as it approaches the subject, while operating at altitudes of 3 m, 4 m, and 5 m, and scanning horizontal distances ranging from 17 m to 2 m in 0.5 m intervals. Drones labelled 1, 2, and 3 represent different positions of the same drone across the capture site. The angle of depression shown in the figure denotes the angle between the drone’s horizontal line of sight and the downward direction towards the subject. \textbf{(b)} The complete workflow of the proposed application. Non-frontal (NF) images captured from the specified drone positions, as shown in \textbf{(a)}, are frontalised by the face reconstruction unit (FRU), and the resultant synthetic frontal (SF) images are classified by the memristor-based face classification unit (FCU). \textbf{(c)} Average identification accuracy per altitude setting on the DroneFace dataset using the FCU, before (baseline) and after FRU reconstruction.}\label{main_figure_3}
  \end{center}
\end{figure}

\subsection{GAN-based frontalisation for mitigating pose variability}

To mitigate the challenges associated with non-frontal pose classification, non-frontal images were reconstructed using the face reconstruction unit (FRU), as illustrated in Figures \ref{main_figure_2}(a) and \ref{main_figure_2}(b). Non-frontal images are input to the generator, where they are encoded into a latent representation and subsequently decoded to synthetic frontal images through progressive upsampling and feature refinement. The generated synthetic frontal images are then passed to the discriminator, which evaluates their authenticity against corresponding ground-truth images. Through this adversarial learning process, the generator learns to produce high-resolution frontal representations that conform to the target domain distribution.

We selected 150 subjects from the CMU Multi-PIE dataset \cite{4813399} for training and a further 24 subjects for validation. After training, the FRU performed inference to reconstruct non-frontal images that were used in the preceding experiment. Details of the FRU training procedure are provided in Supplementary Tables S2 and S3.

Figure \ref{main_figure_2}(c) presents the generation results for seen subjects across different camera angles. For each subject, the top row shows the synthetic frontal (SF) images, while the bottom row shows the corresponding non-frontal images. Ground-truth (GT) images are also provided for reference.

Figure \ref{main_figure_2}(d) shows the average identification accuracy of synthetic frontal images for each camera angle bin, defined based on the corresponding non-frontal input angles. The identification results for the reconstructed images demonstrate a substantial improvement compared to the baseline non-frontal results across all pose angles. The proposed approach achieves consistently higher accuracy across all bins, with mean values of 96.31\% ± 2.53\% at ±15°, 96.24\% ± 2.49\% at ±30°, 96.29\% ± 2.31\% at ±45°, 96.49\% ± 2.51\% at ±60°, 96.22\% ± 2.07\% at ±75°, and 96.16\% ± 2.60\% at ±90°. In contrast to the baseline, which exhibited a monotonic degradation in accuracy from 35.13\% at 15° to 7.44\% at 90°, the proposed method maintains a stable and high-performance profile across all pose variations. Supplementary Figure S11 presents confusion matrices for the corresponding experiment across all camera-angle bin conditions. Additionally, we conducted further experiments on how memristor resistance tolerance impacts identification accuracy. The results are presented in Supplementary Figure S8, which shows the training and test accuracies, and Supplementary Figure S9, which presents confusion matrices for experiments run with different resistance tolerance values. These results demonstrate the effectiveness of the proposed GAN-based FRU in improving identification performance. The method mitigates pose-dependent degradation observed in the baseline while maintaining high accuracy across all viewing angles. Minor fluctuations across angle bins are attributed to the stochastic nature of GAN-based generation.

\subsection{Beyond controlled environments: edge-based evaluation}

Up to this point, we evaluated the proposed facial recognition framework on the CMU Multi-PIE dataset \cite{4813399}, which contains a large number of images per subject acquired under controlled conditions while still exhibiting substantial variability. However, in edge deployment scenarios, application-specific constraints such as high dynamism and significant mobility must also be considered. To evaluate the proposed system under such scenarios, we benchmarked the framework on the DroneFace dataset \cite{10.1145/3083187.3083214}, which reflects real-world operating conditions.

DroneFace consists of 1,364 images of 11 subjects captured across varying ground distances (2–17 m in 0.5 m increments) and altitudes (1.5, 3, 4, and 5 m). Images at 1.5 m were acquired using a fixed camera, whereas those at higher altitudes were captured using a drone-mounted camera. Figure \ref{main_figure_3}(a) illustrates the angle-of-depression map across altitude-distance combinations, where the angle of depression is defined as the angle between the drone’s horizontal axis and its line of sight to the subject. Additional details on the dataset image-size distribution and viewing geometries are provided in Supplementary Figure S12. Given the limited dataset size, we applied a range of image augmentation techniques to improve diversity and evaluation robustness. Details of the augmentation procedures are provided in Supplementary Figure S13 and Supplementary Note 1.

To effectively utilise the dataset and ensure a coherent evaluation, we trained the FCU using images captured at a height of 1.5 m, along with the original ground-truth (GT) portraits of the 11 subjects. Inference was then performed on images from the remaining height settings (3 m, 4 m, and 5 m) to assess the effect of varying angles of depression on identification performance. The experiment was performed on baseline (pre-FRU) images and repeated using FRU-generated synthetic frontal images. Details of the FRU training for this experiment are provided in Supplementary Tables \addtext{S5 and S6}. Figure \ref{main_figure_3}(b) illustrates the complete workflow of the proposed facial classification application, showing high-level representations of both the FRU and FCU. Figure \ref{main_figure_3}(c) presents the average identification accuracies for baseline and post-FRU results. Supplementary Figure S14 illustrates key aspects of the memristor-based training process for this experiment, including the evolution of device memristive states over the training session, the pre- and post-training conductance distributions, and the spike activity history. Further training details of the FCU for this experiment are presented in Supplementary Table S4.

For non-frontal baseline images, the average identification accuracy decreases from 69.85\% ± 7.21\% at 3 m to 50.62\% ± 5.44\% at 4 m and 40.47\% ± 5.98\% at 5 m, indicating a degradation in performance with increasing altitude. In contrast, the reconstructed synthetic frontal images achieve substantial improvements over the baseline, with up to 82\% higher identification accuracy. Specifically, the mean accuracies are 73.30\% ± 7.71\%, 73.35\% ± 7.40\%, and 73.60\% ± 7.07\% at 3 m, 4 m, and 5 m, respectively, compared to non-frontal baseline images. While the gains are modest at lower altitudes, where baseline performance is already relatively high, the improvement becomes substantial at higher altitudes, demonstrating the effectiveness of frontal reconstruction in mitigating severe pose variations. However, the elevated standard deviation in the reconstructed results indicates increased variability, likely due to intrinsic variability in the GAN model, which may affect output consistency.

Supplementary Figure S15 presents confusion matrices evaluating the corresponding experiment across all altitude settings, for both pre- and post-FRU identification cases.

\subsection{Handling unseen identities in open-set settings}

\begin{figure}
  \begin{center}
  \includegraphics[width=\linewidth]{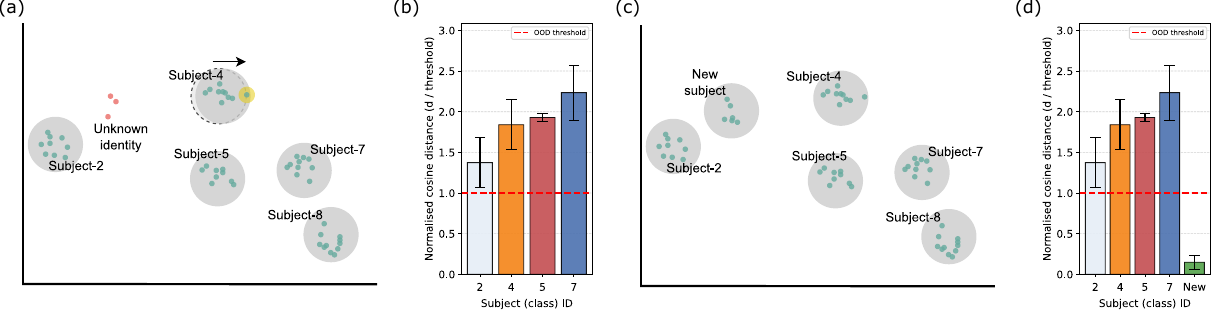}\\
  \caption{\textbf{Prototype-based incremental learning and novel class registration.} \textbf{(a)} A labelled test sample from Subject-4 appears at the cluster periphery, and the class prototype shifts dynamically towards the sample in the embedding space. Additionally, input samples from an out-of-distribution (OOD) identity are streamed and embedded away from existing class clusters, gradually forming a new cluster labelled as "Unknown identity". \textbf{(b)} Average normalised cosine distance of the last three arrived samples from the "Unknown identity", later recognised as "New subject" in (c), relative to existing subject clusters. The mean and standard deviation are computed across these samples and normalised by the decision threshold.  \textbf{(c)} Once the cluster reaches sufficient consistency, the identity is incorporated into the in-distribution set by registering it as "New subject". \textbf{(d)} Corresponding distances after adaptation following registration of the new subject.}\label{main_figure_4}
  \end{center}
\end{figure}

In the previous section, we demonstrated that the proposed face recognition framework is effective in mobile and resource-constrained environments such as aerial surveillance. The experiments thus far assume a closed-set setting, where inference is performed via 1:\textit{N} matching and all test identities are known \textit{a priori}. While computationally efficient and suitable for applications with fixed identity sets, such as access control and watchlist-based law enforcement, this assumption rarely holds in unconstrained environments. This limitation motivates open-set recognition approaches \cite{9040673, 9304882, Gnther2017TowardOF}, which incorporate explicit unknown detection alongside standard identification.

Beyond unknown rejection, real-world deployments also require mechanisms to retain and organise observations of previously unseen individuals. This motivates online or incremental learning, enabling continuous updates to identity representations as new data become available. For instance, in dynamic security settings, individuals initially observed as unknown may be progressively added to a watchlist based on repeated appearances or behavioural cues, while in large-scale operations, unregistered persons may later be included in monitoring lists following repeated detections or investigative feedback. Therefore, we implemented an auxiliary prototype-based incremental learning scheme, enabling continuous adaptation of class prototypes during inference without retraining the FCU or causing catastrophic forgetting \cite{MCCLOSKEY1989109}.

The adaptive learning framework was evaluated using the same FCU configuration employed in the DroneFace experiments, with a fixed manual seed of 201. Initial prototypes were generated for all 11 registered subjects using embeddings extracted from the DroneFace training set. During inference, 121 FRU-reconstructed frontalised test images captured at an altitude of 5 m were streamed to the system to evaluate online adaptation. To assess incremental identity registration, 7 frontalised images from the CMU Multi-PIE dataset (ID 028), originally captured at a 15° camera angle, were additionally introduced as samples from previously unseen identities.

As labelled samples from known subjects are streamed, their corresponding prototypes progressively shift towards the incoming embeddings through online updates. Figure \ref{main_figure_4}(a) shows the t-distributed stochastic neighbour embedding (t-SNE) representation of the known subject classes, including an example test sample located near the periphery of the Subject-4 prototype boundary. The dashed circle denotes the pre-update prototype boundary, highlighting the adaptive displacement of the class representation following online updating.

The framework further demonstrates robust handling of previously unseen identities through dynamic prototype registration. Figure \ref{main_figure_4}(a) also shows samples from an unknown identity, initially rejected because of their separation from existing class clusters. As additional samples are observed, they form an increasingly coherent grouping within the embedding space. Figure \ref{main_figure_4}(b) shows the average distances between three representative samples and all existing prototypes before registration. Once the buffered embeddings satisfy the registration criteria, a new prototype corresponding to the previously unseen identity is instantiated, as shown in Figure \ref{main_figure_4}(c). Figure \ref{main_figure_4}(d) further shows the average distances between the same samples and all prototypes after registration.

\section{Methods}\label{sec3}

\subsection{Device characterisation and model fitting}

TiN/HfON/TiN devices were characterised on the ArC ONE\texttrademark{} platform \cite{7113814} to construct an empirical model for FCU experiments. We first performed pulse-based electroforming on pristine devices to activate resistive switching via conductive filament formation. Electroforming was carried out using voltage pulses with widths of 10--100~$\mu$s and amplitudes of 2.5--5.5~V. Following electroforming, current--voltage (I--V) measurements were performed using $\pm$2~V sweeps with 2~ms voltage pulses under defined sweep rates and compliance limits. Device resistance was recorded during each sweep. Devices exhibiting non-linear I--V behaviour and hysteresis were subjected to further experiments to identify the voltage range and polarity associated with resistive switching. This allowed definition of bias conditions for subsequent model fitting. The process consists of two parts. First, alternating-polarity voltage pulses with progressively increasing amplitudes were applied to determine polarity-dependent switching. Second, pulsed voltage ramps from 0.5 to 3.5~V were used to induce switching, using 100~$\mu$s pulse widths and 0.2~V increments, with a 1~M$\Omega$ resistance threshold. We used the voltage window corresponding to significant resistance modulation for parameter extraction. Supplementary Figure S1(a) illustrates the experimental setup used for the characterisation process, while Supplementary Figures S1(c) and (d) show the I--V curve and the switching dynamics of the DUT, respectively. Supplementary Figures S2, S3, S4, S5, and S6 present screenshots from the ArC ONE\texttrademark{} control interface corresponding to different stages of the characterisation process. Details about the switching-rate equation and the definitions of the extracted parameters are provided in Supplementary Note 3.

\subsection{Face frontalisation}

The FRU is built on an image-to-image GAN architecture to reconstruct non-frontal facial images into frontal views. Specifically, as the FRU is expected to generate frontal images from non-frontal profile inputs, its generator does not initially rely on a random noise vector, as in conventional GANs. Instead, it receives the profile images directly to produce corresponding frontal views. To achieve this, the generator follows an encoder–decoder design \cite{WANG201937, 7803544} with U-Net-style skip connections \cite{DBLP:journals/corr/RonnebergerFB15}, whereas the discriminator is based on the DCGAN-inspired \cite{radford2016unsupervised} convolutional architecture, with minor modifications for $128\times128$ pixel images. The FRU was implemented in PyTorch \cite{paszke2019pytorchimperativestylehighperformance}.

The complete generator architecture is illustrated in Figure \ref{main_figure_2}(a). As shown in the figure, non-frontal images of size $128\times128$ pixels are fed into the generator’s encoder. This part consists of a series of convolutional layer blocks, some of which contain additional components such as batch normalisation and LeakyReLU. The generator depth parameter ($ngf$) controls the number of features extracted at each stage and can be adjusted depending on computational constraints. Once features pass through the encoder and the spatial resolution reaches $2 \times 2$, the encoded representations are routed through a bottleneck and into the decoder. The decoder reconstructs frontal views using transposed convolutions, mirroring parts of the encoder’s structure. Skip connections from the encoder to selected decoder layers allow feature maps at matching resolutions to be concatenated, preserving fine-grained spatial information otherwise lost during downsampling. This enables the decoder to combine high-level semantic features from the bottleneck with spatial details from earlier layers. As decoding proceeds, spatial resolution increases while the number of feature maps decreases, ultimately producing the final output image. A Tanh activation generates the normalised synthetic frontal image, which is then passed to the discriminator for evaluation. The generator is optimised through adversarial learning against the discriminator, using loss functions that quantify the discrepancy between generated and real frontal images. The specific formulation of the generator loss functions, which define the training objective for image reconstruction and adversarial consistency, is provided in Supplementary Note 2.

The discriminator processes both ground-truth and synthetic frontal images through a series of convolutional layers with LeakyReLU activations and batch normalisation. The network progressively reduces spatial resolution while increasing the number of feature maps, producing a final $4 \times 4 \times 1$ map of real/fake logits. The discriminator is trained using binary cross-entropy loss with logits. The number of feature maps ($ndf$) is a configurable hyperparameter, analogous to $ngf$ in the generator.

\subsection{Face classification}

The FCU was implemented using the NeuroPack simulator \cite{Huang_2022, 10666111}, which we further extended to support three-channel RGB inputs and higher-complexity facial imagery. 

The original NeuroPack simulator \cite{Huang_2022} required three essential input files: a configuration file defining network, neuron, and memristor parameters; a connectivity matrix specifying synaptic mappings to the virtual memristor crossbar array; and stimuli files containing spike-based training and test data generated from image inputs.

The feature extraction core consists of convolutional layers followed by a fully connected layer that generates compact visual representations to support memristor-based classification. After feature extraction, the resulting feature maps are converted into spike-based representations according to the selected encoding scheme, from which training and test stimuli files are generated. These spike trains are processed by the virtual memristor crossbar array, whose behaviour is defined through configuration files specifying network, neuron, and device-level parameters, together with a connectivity matrix encoding synaptic mappings and network topology. Each entry in the connectivity matrix defines the pre- and post-synaptic neuron indices, corresponding crossbar coordinates, and synaptic connection type \cite{6572171}.

The crossbar size is determined by the number of extracted features mapped onto the array. For the FCU, input images of size $128 \times 128 \times 3$ are reduced to 512 features per class, yielding 5,632 total features for an \addtext{eleven-class} dataset and corresponding to a $76 \times 76$ crossbar topology. The memristor crossbar fully connected layer forms the computational core of the simulator, where spike-driven neural computation and synaptic plasticity are implemented.

The neuron core computes neuronal dynamics from the resistive states of the virtual memristor crossbar under spike-based inputs, including membrane potential evolution. A winner-take-all (WTA) mechanism is applied at the network level, such that only the neuron with the highest membrane potential is activated when multiple neurons exceed their firing threshold, enforcing a sparse decision rule analogous to a softmax-based selection \cite{nwankpa2018activationfunctionscomparisontrends}. The resulting firing events are accumulated and passed to the plasticity core, which updates synaptic weights according to the selected learning rule during training. During inference, the output is determined by the winning neuron.

\addtext{Benchmark results for different spike encoding schemes and numbers of time steps $T$ are presented in Supplementary Table S7. Further details on the algorithms underlying \addtext{the} FCU are provided in Supplementary Note 4.}

\subsection{Adaptive learning}

To enable continual adaptation during deployment, we introduced an online prototype-learning framework that operates on spike-derived output embeddings generated by the FCU. Neuronal activity is accumulated over a fixed spike-train window, and the resulting output-layer response vector is treated as a compact embedding in the network’s latent space. Class prototypes are initialised from the training set as class-wise centroids of these embeddings.

During inference, samples are assigned via nearest-prototype matching using configurable distance metrics, including cosine, Euclidean and Mahalanobis distances. Open-set recognition is enabled through a fixed rejection threshold, whereby embeddings outside the prototype boundary are classified as unknown, allowing distinction between known and unseen identities without retraining the FCU.

Online adaptation is performed using exponential moving average updates, enabling gradual refinement of prototypes while preserving previously learned structure. For labelled samples, updates are applied with confidence-weighted learning rates derived from agreement between prototype-based prediction and the WTA output of the spiking network. For unlabelled samples from known classes, prototype adaptation is permitted only when the prototype assignment \addtext{agrees} with the network’s WTA output, providing a conservative gating mechanism that suppresses unreliable updates and mitigates catastrophic drift during continual learning.

To support emergence of novel identities, rejected embeddings are accumulated in a buffer and monitored for consistency over time. When sufficient evidence of a new identity is observed under predefined confidence and buffer-size criteria, a new class prototype is instantiated by averaging the buffered embeddings and assigning a unique label.

Supplementary Figure S16 illustrates the flowchart of the adaptive learning process, with further details on its underlying mechanisms provided in Supplementary Note 5.

\section{Conclusions}\label{sec4}

In this work, we propose a complementary facial recognition framework that combines a lightweight GAN-based face reconstruction unit (FRU) with a resource-aware memristor-based classification architecture (i.e., the face classification unit, FCU). The approach addresses degraded recognition performance under non-frontal viewpoints, particularly in compact systems where computational constraints preclude the use of resource-intensive models. To balance early-stage hardware design with practical inference evaluation, we integrate architectural simulation with empirical modelling of the classification process, enabling hardware-representative assessment while maintaining algorithmic flexibility.

Experiments on two datasets demonstrate that pose variation remains a dominant factor affecting identification accuracy, with severe degradation observed under extreme yaw (horizontal head rotations), where performance drops from 35.13\% at ±15° to 7.44\% at ±90°. The proposed reconstruction stage mitigates this effect, substantially improving recognition on frontalised representations, reaching up to 96\% accuracy under the corresponding settings. Improvements are also observed under pitch (vertical head rotations) variations, where performance under elevated conditions remains above 40\% at baseline and increases to an average of 73.4\% after reconstruction. Overall, the results indicate that compact generative reconstruction combined with memristor-based classifiers enhances robustness under both constrained and dynamic imaging conditions. \addtext{Among the existing GAN-based face frontalisation and memristor-based recognition methods in the literature, which are summarised in Supplementary Table S8 along with their key insights, we position our work as a bridge between compact generative models and efficient neuromorphic algorithms. By integrating these two complementary directions, our work helps close the gap between face frontalisation and resource-efficient recognition, paving the way towards the practical deployment of an end-to-end system for resource-constrained edge applications.}

The results on the DroneFace dataset also highlight the challenge of maintaining identification accuracy at greater drone altitudes, where reduced facial resolution and changes in viewing geometry limit the availability of discriminative facial information. \addtext{This challenge is compounded by differences between the two datasets themselves: unlike Multi-PIE, which is captured under controlled studio conditions, DroneFace reflects real-world, uncontrolled capture with a drone-mounted camera at varying distances and altitudes, and its overall lower accuracy relative to Multi-PIE reflects the combined effect of a substantially smaller dataset, lower baseline resolution, and less constrained acquisition conditions.} Although the super-resolution preprocessing employed in this work provides a complementary mechanism for enhancing facial detail prior to GAN-based frontalisation, future work could investigate identity-preserving super-resolution and joint restoration-recognition learning \cite{ataercansizoglu2019verificationlowresolutionfacesusing, XU2021158}, together with training across a broader range of altitudes and image resolutions using more comprehensive and diverse, dedicated drone-captured face recognition datasets. Furthermore, multi-scale feature learning, including approaches based on pyramid pooling and atrous spatial pyramid pooling, together with temporal fusion of consecutive drone observations, represents promising directions for improving recognition robustness under challenging high-altitude conditions \cite{MISHRA2021104290, 8100143, chen2017deeplabsemanticimagesegmentation, 4401949}.

Beyond the core WTA-based classification mechanism, the proposed prototype-based secondary module enables the incorporation of previously unseen subjects, treating them as in-distribution samples for future reference, thereby demonstrating potential for incremental system adaptability.

\addtext{The memristor-based part of this work is performed at the simulation level. Validation on physical crossbar arrays, including hardware non-idealities beyond those modelled here (e.g., device failure and conductance drift), and investigating the feasibility of deploying the GAN-based FRU on memristive hardware \cite{Krestinskaya2020} remain future work.}

This work has centred on memristive devices as a computational substrate for efficient, edge-compatible facial recognition. It is worth noting, however, that memristive technology is not limited to this role: the same non-linear, history-dependent dynamics that enable neuromorphic classification have separately been exploited to generate lightweight, resource-efficient encryption keystreams for image and video data \cite{11142957}. This points to a broader design space in which memristive devices of the kind used here for onboard classification could, in future work, also support encrypted data transmission by exploiting these same non-linear dynamics. For high-stakes deployments, where confidentiality of captured facial imagery is critical, this dual-purpose potential offers a route towards unified, resource-constrained pipelines.

\section*{Acknowledgments}

This work was supported by the Engineering and Physical Sciences Research Council (EPSRC) AI Hub for Productive Research and Innovation in eLectronics (APRIL) under Grant No. EP/Y029763/1, and by the Royal Academy of Engineering (RAEng) Chair in Emerging Technologies under Grant No. CiET1819/2/93.

\section*{Competing interests}

The authors declare no conflict of interest.


\bigskip





\clearpage
\bibliography{sn-bibliography}

\clearpage

\input{supplementary}

\end{document}

%% file: supplementary.tex

\setcounter{page}{1}
\renewcommand{\thepage}{\arabic{page}}


\thispagestyle{empty}

\vspace*{2.5cm}

\begin{center}

{\LARGE
Non-frontal face recognition using GANs\\
and memristor-based classifiers\\
\bfseries Supplementary Information
\par}

\vspace{1.5em}

{\large
Semih Vazgecen\textsuperscript{*}, Cristian Sestito,
Spyros Stathopoulos and Themis Prodromakis
\par}

\vspace{1em}

{\small
Centre for Electronics Frontiers, Institute for Integrated Micro and Nano Systems,\\
School of Engineering, The University of Edinburgh, UK\\[0.2in]
\textsuperscript{*}Corresponding author. E-mail:
S.A.Vazgecen@sms.ed.ac.uk;
\par}

\vspace{1em}

\end{center}


\renewcommand{\thefigure}{S\arabic{figure}}
\setcounter{figure}{0}

\renewcommand{\thetable}{S\arabic{table}}
\setcounter{table}{0}


\begin{bibunit}

\clearpage

\begin{figure}
  \begin{center}
  \includegraphics[width=\linewidth]{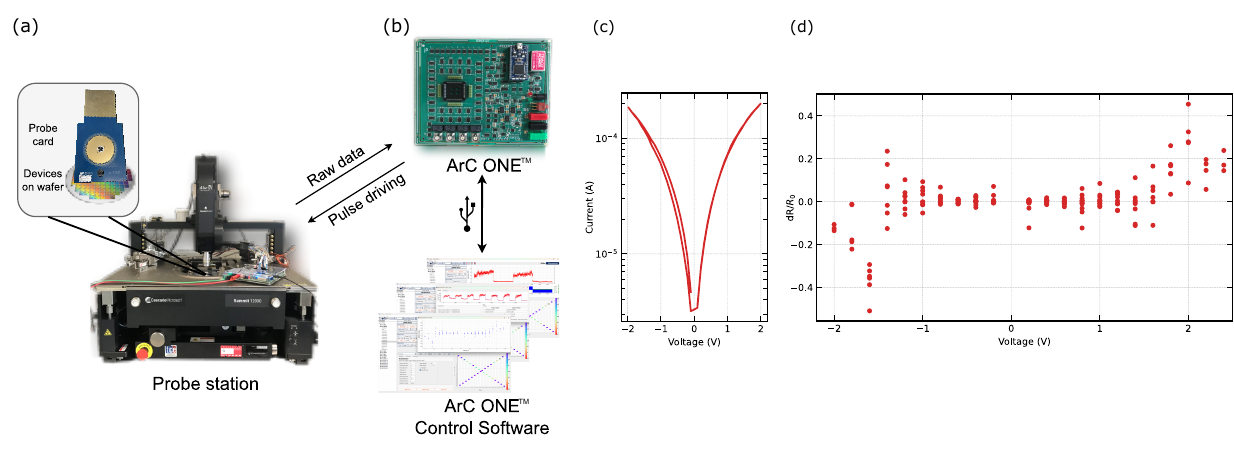}\\
  \caption{\textbf{Experimental setup for empirical model creation and further characterisation results.} \textbf{(a)} The probe station was used during the process along with the probe card and a representational image for devices on wafer. \textbf{(b)} the ArC ONE\texttrademark{} characterisation board and screenshots from its control interface. \textbf{(c)} I--V curve of the electroformed TiN/HfON/TiN device used to create the empirical model. \textbf{(d)} Switching dynamics of the corresponding device, illustrating the relative resistance change induced by a programming voltage, referenced to the initial resistance ($R_0$).}\label{S1}
  \end{center}
\end{figure}

\begin{figure}
  \begin{center}
  \includegraphics[width=\linewidth]{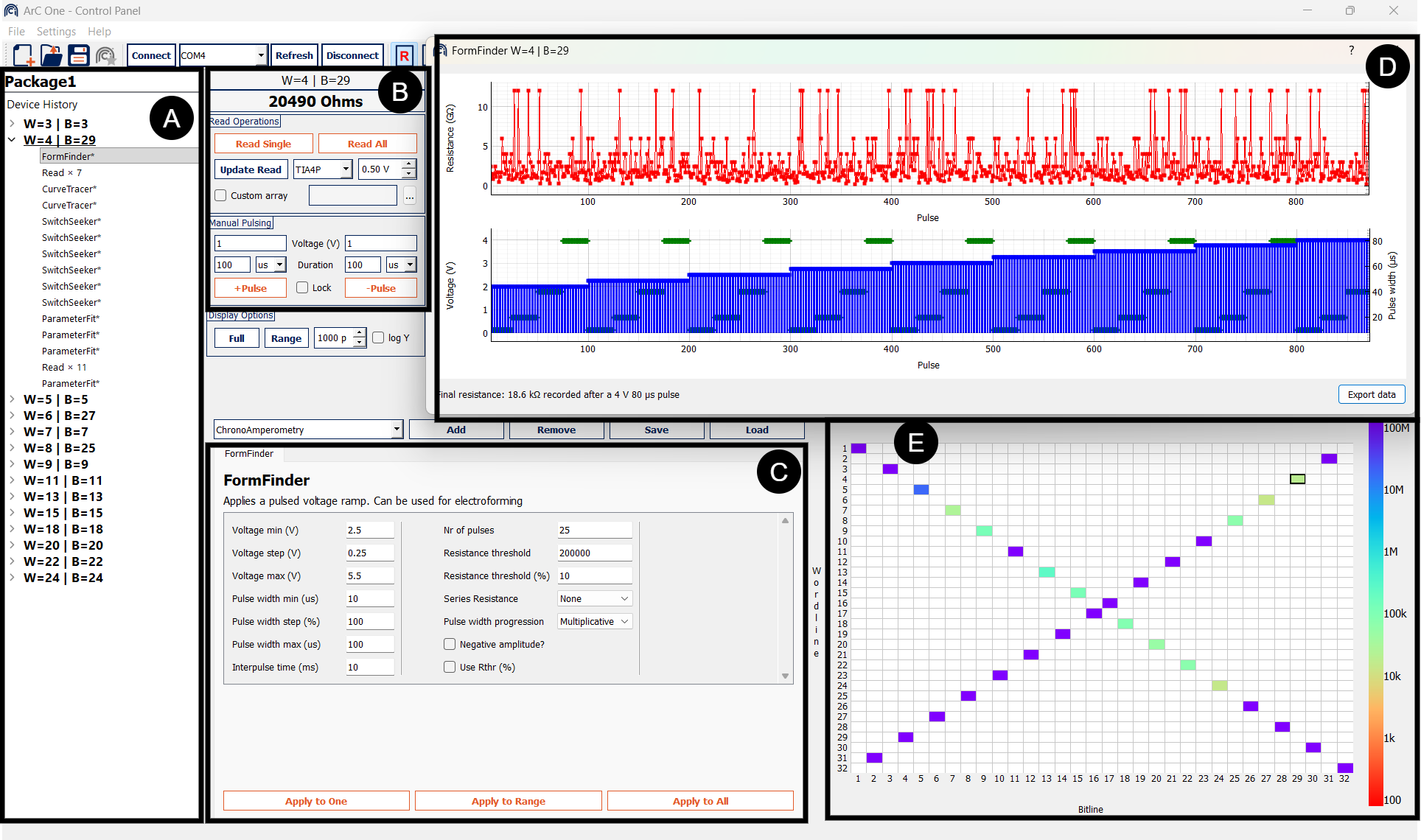}\\
  \caption{\textbf{Annotated screenshot from the ArC ONE\texttrademark{} Control Software.} Panel A: device history list. Panel B: current resistance status of a selected device and manual read/pulse operation controls. Panel C: control panel for selected dedicated biasing modules. Panel D: data plot based on the applied biasing scheme. Panel E: crossbar map.}\label{S2}
  \end{center}
\end{figure}

\clearpage

\begin{figure}
  \begin{center}
  \includegraphics[width=\linewidth]{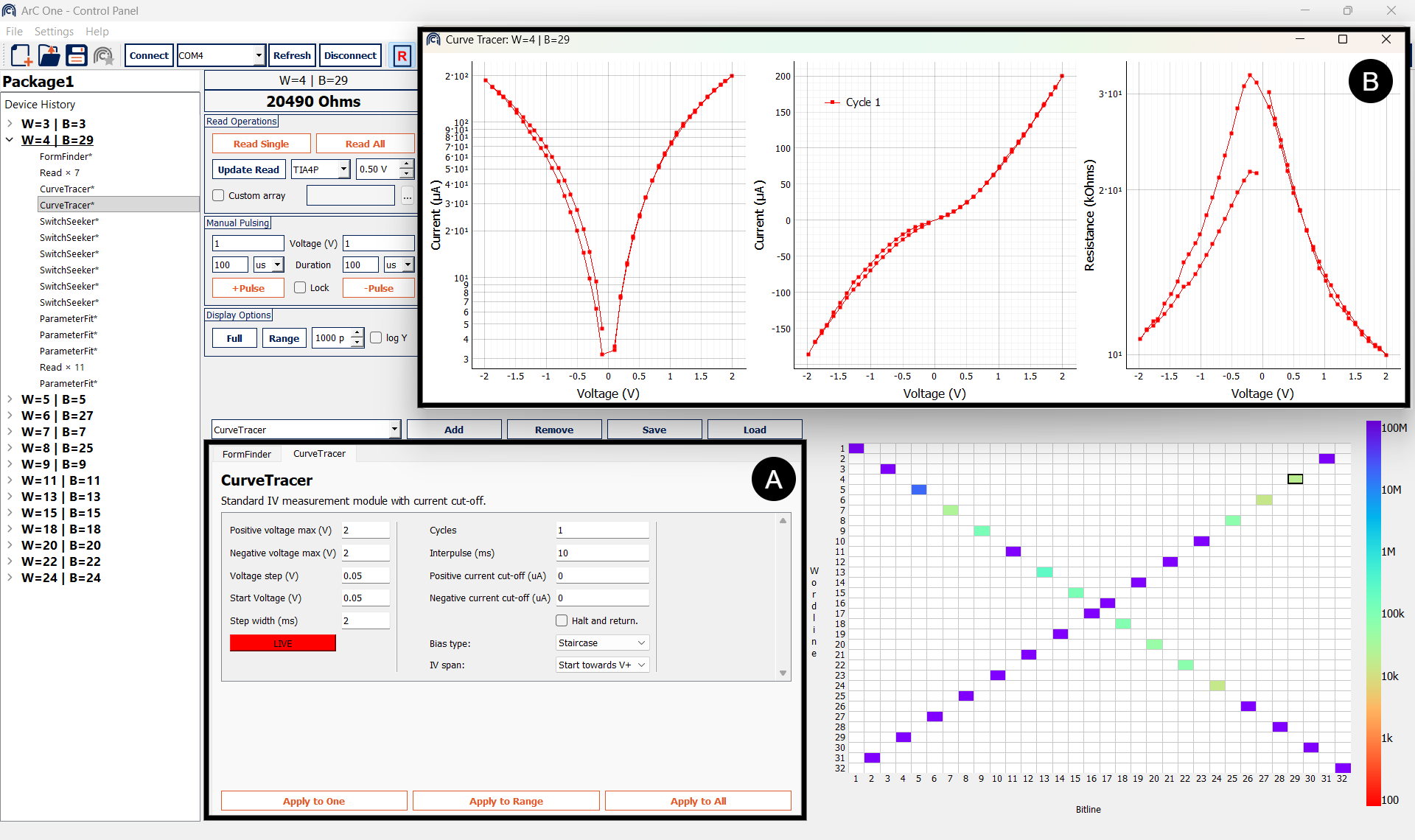}\\
    \caption{\textbf{Screenshot from the I--V characteristics extraction process.} Panel A: module settings. Panel B: experimental results, showing (from left to right) the I--V curve in logarithmic scale, the I--V curve in linear scale, and the R–V characteristics in logarithmic scale.}\label{S3}
  \end{center}
\end{figure}

\begin{figure}
  \begin{center}
  \includegraphics[width=\linewidth]{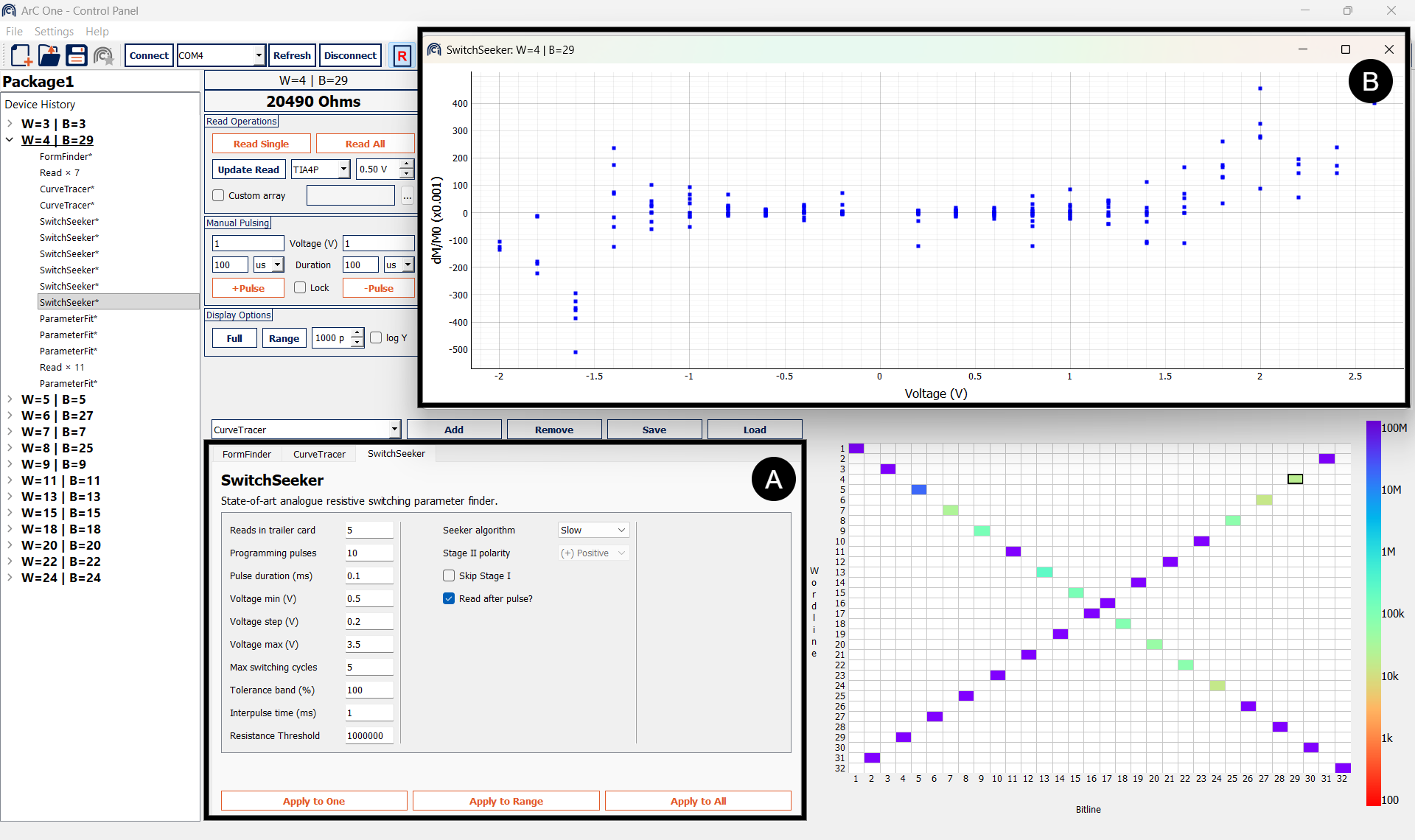}\\
    \caption{\textbf{Screenshot illustrating the device switching dynamics analysis process.} Panel A: module settings. Panel B: experimental results showing relative resistance changes across varying voltage levels.}\label{S4}
  \end{center}
\end{figure}

\clearpage

\begin{figure}
  \begin{center}
  \includegraphics[width=\linewidth]{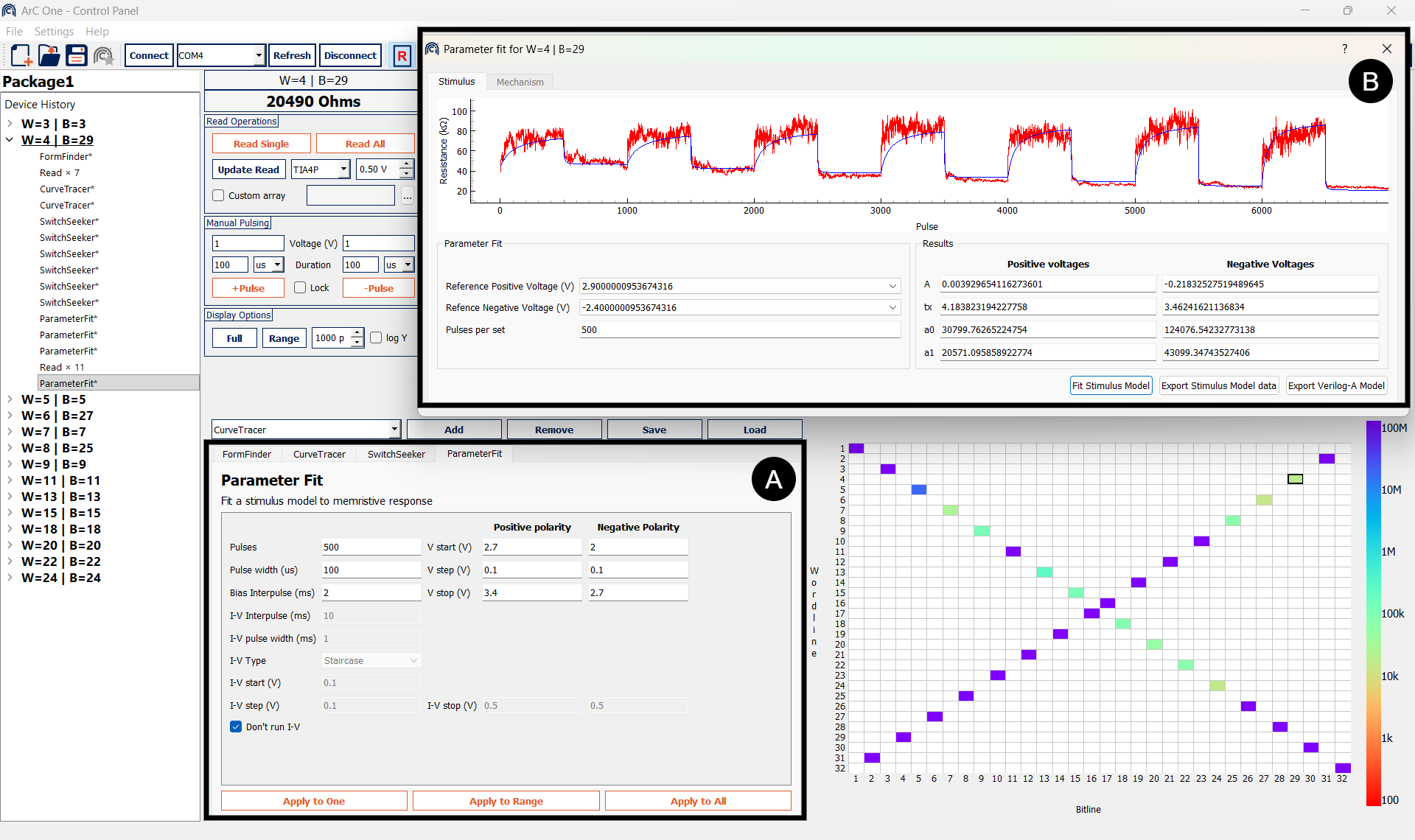}\\
    \caption{\textbf{Screenshot of the parameter fitting procedure.} Panel A: module settings. Panel B: fitting results along with extracted parameters.}\label{S5}
  \end{center}
\end{figure}

\begin{figure}
  \begin{center}
  \includegraphics[width=\linewidth]{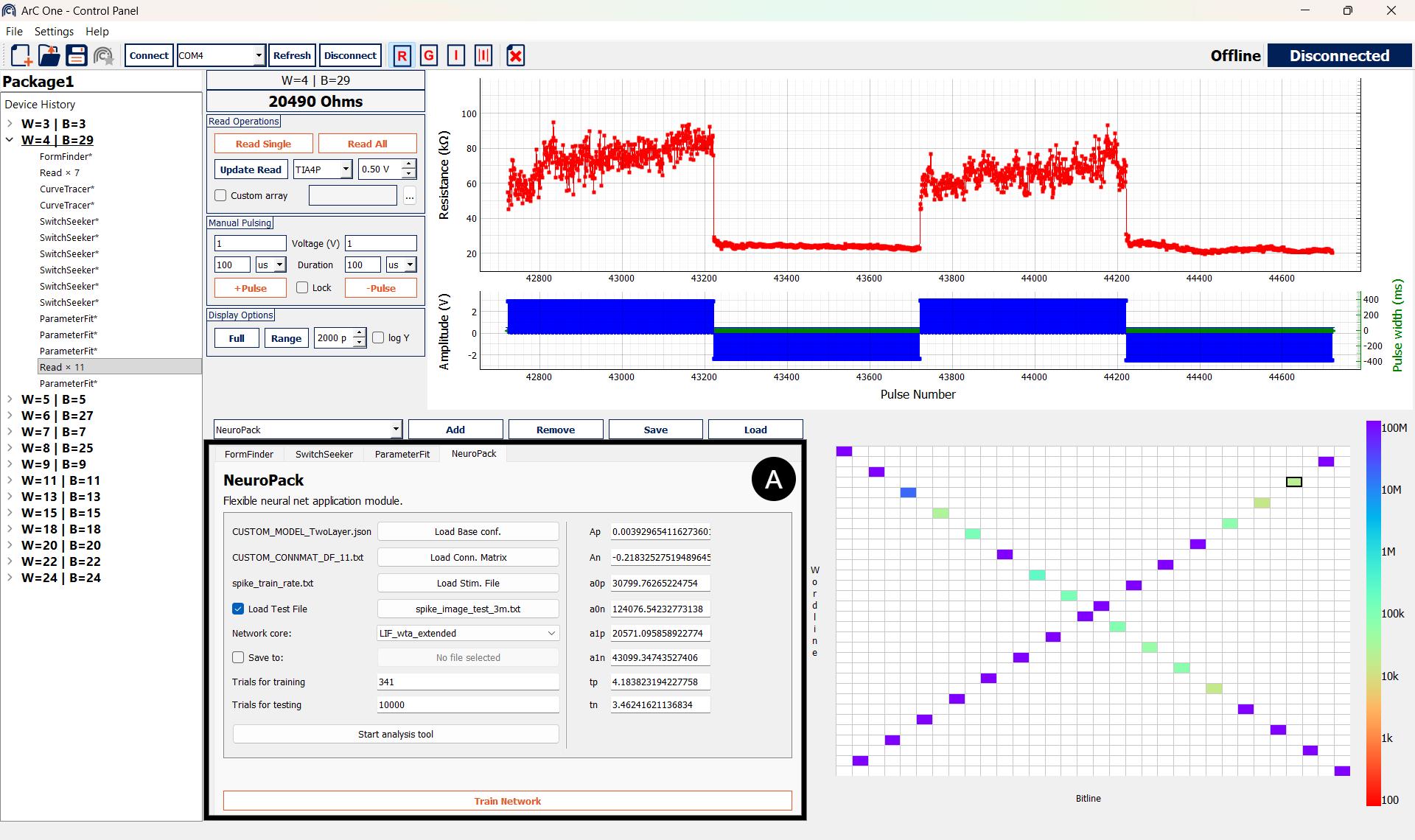}\\
    \caption{\textbf{Screenshot of the memristor simulator module, NeuroPack.} The module takes various input file types along with memristor model parameters to run simulations.}\label{S6}
  \end{center}
\end{figure}

\begin{figure}
  \begin{center}  \includegraphics[width=\linewidth]{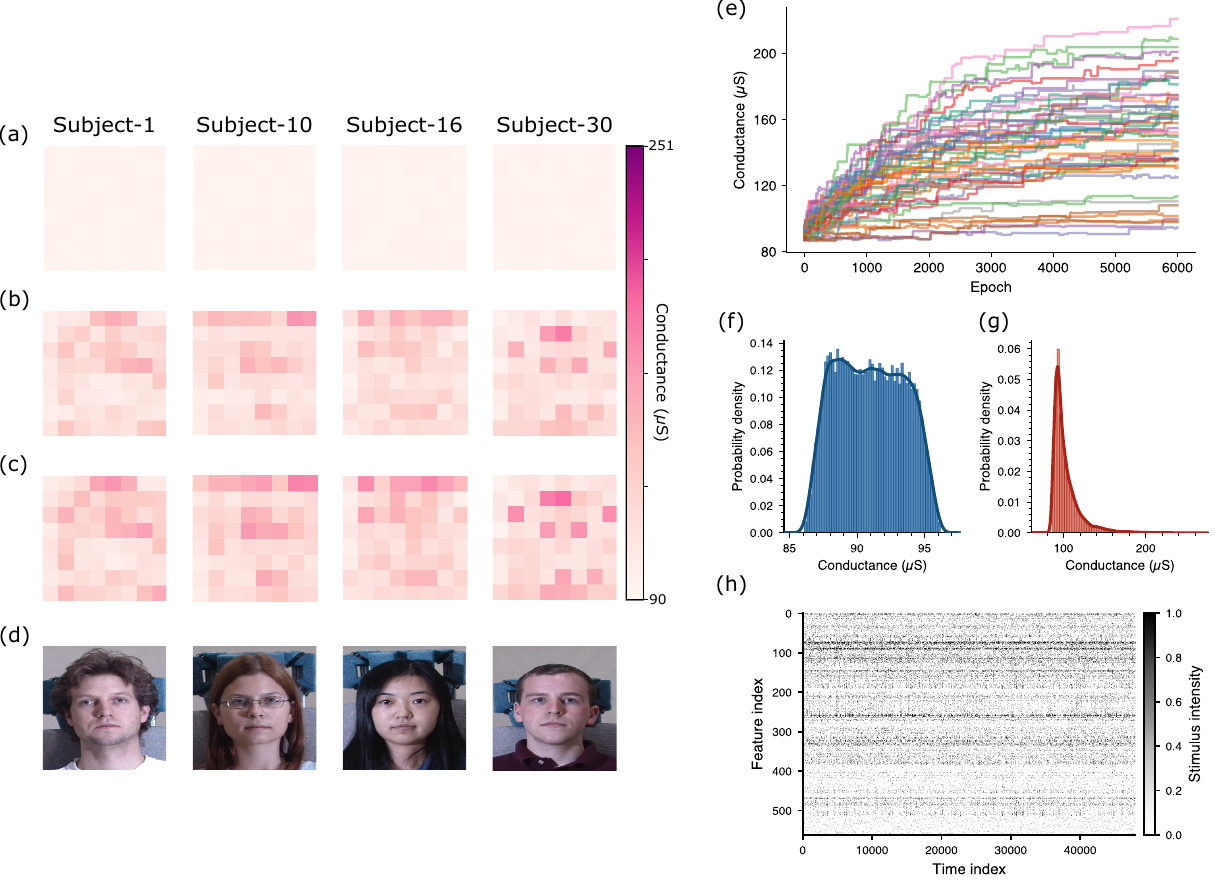}\\
  \caption{\textbf{Memristor-based classification dynamics on the CMU Multi-PIE dataset \cite{supp-4813399}.} \textbf{(a)} Memristor resistive states for the given subject classes before training. \textbf{(b)} Memristor resistive states at 50\% of training. \textbf{(c)} Memristor resistive states after training. \textbf{(d)} Representative ground-truth images of the corresponding subjects. \textbf{(e)} Learning curve of a synapse across all subject classes during training. \textbf{(f)} Initial conductance distribution before training. \textbf{(g)} Post-training conductance distribution. \textbf{(h)} Raster plot showing neuronal spike activity over time.}
    \label{S7}
  \end{center}
\end{figure}
\clearpage
\begin{figure}
  \begin{center}
  \includegraphics[width=0.92\linewidth]{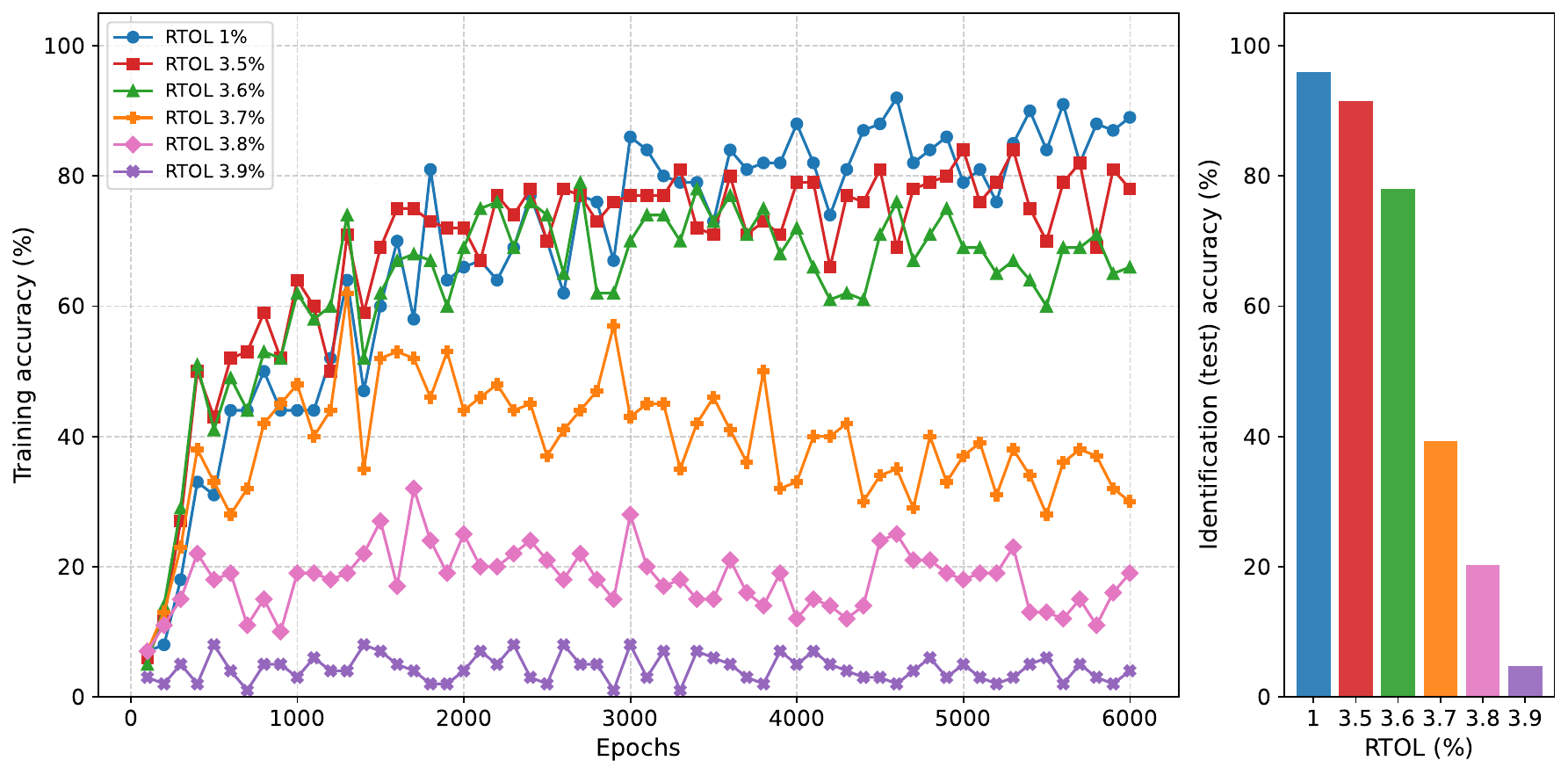}\\
    \caption{\textbf{Investigation of the impact of R-tolerance on identification accuracy.} The figure illustrates classification accuracy for different R-tolerance (RTOL) values. R-tolerance is defined as the maximum allowable relative error between the programmed and target resistance, used as the convergence criterion for the write–verify update process. Iterative updates terminate when this error falls below the R-tolerance or when the maximum number of programming steps is reached. All experiments were conducted using a random seed of 201. Training was performed on the CMU Multi-PIE dataset \cite{supp-4813399}, while inference was conducted on FRU-reconstructed synthetic frontal images originally captured at camera angles of ±30°. Training curves show accuracy recorded every 100 epochs for different RTOL settings, and the bar chart on the right presents the corresponding final identification accuracies.}
    \label{S8}
  \end{center}
\end{figure}
\begin{figure}
  \begin{center}
  \includegraphics[width=0.92\linewidth]{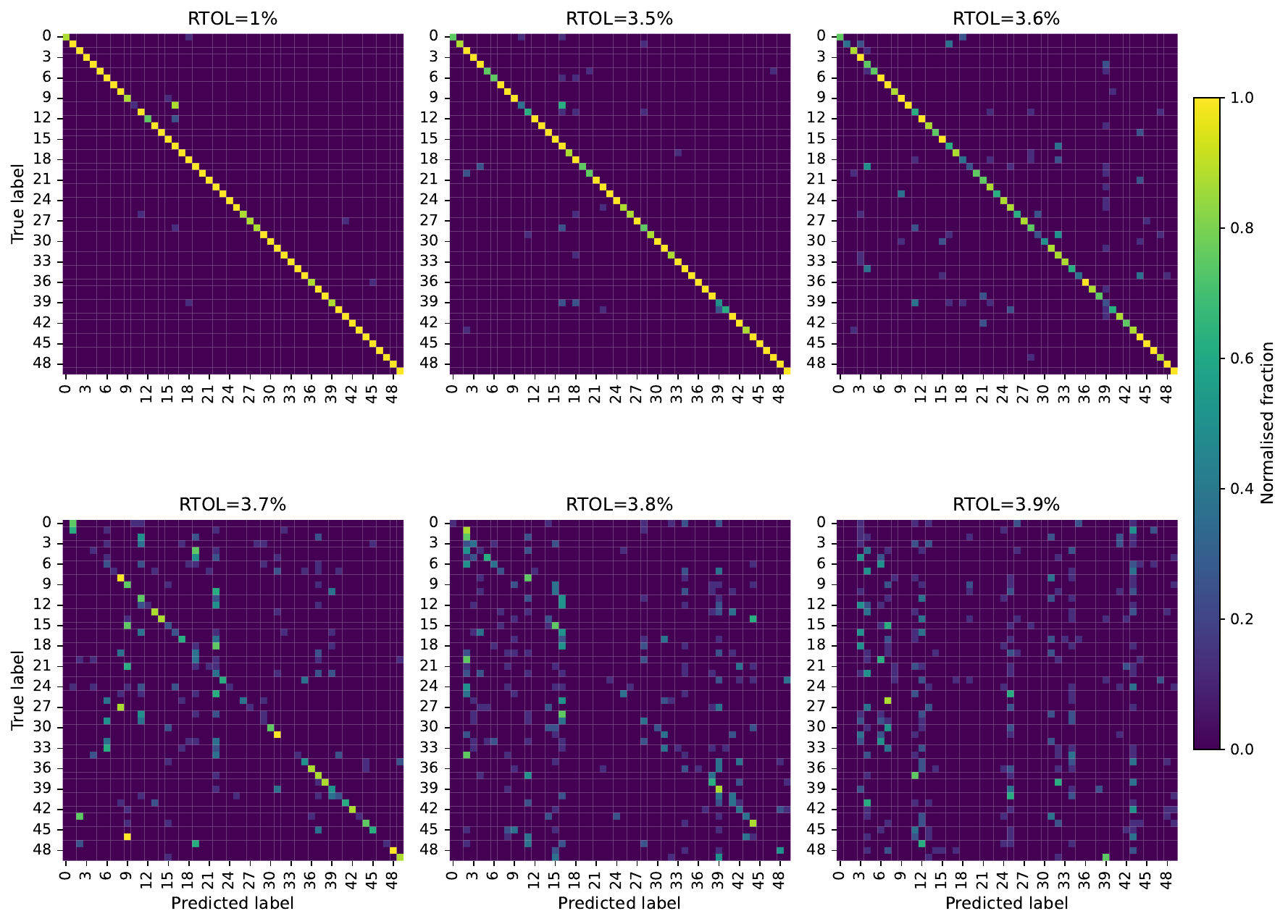}\\
    \caption{\textbf{Confusion matrices corresponding to different R-tolerance (RTOL) settings, as presented in Fig.~\ref{S8}.}}\label{S9}
  \end{center}
\end{figure}
\clearpage
\begin{figure}
  \begin{center}
  \includegraphics[width=\linewidth]{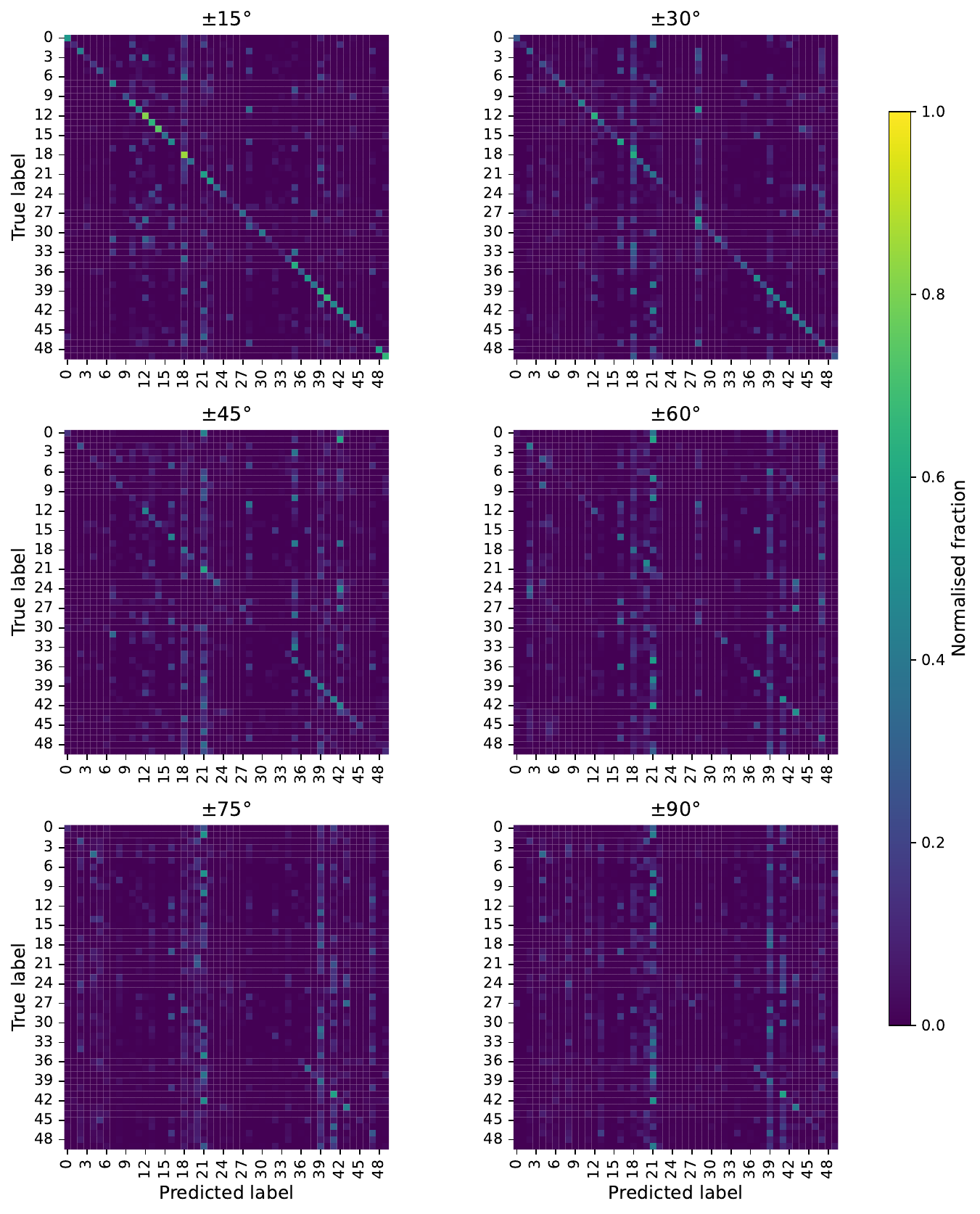}\\
    \caption{\textbf{Confusion matrices for each camera-angle bin in the CMU Multi-PIE dataset \cite{supp-4813399}, corresponding to the pre-FRU classification results reported in Fig. 1.}}\label{S10}
  \end{center}
\end{figure}
\clearpage
\begin{figure}
  \begin{center}
  \includegraphics[width=\linewidth]{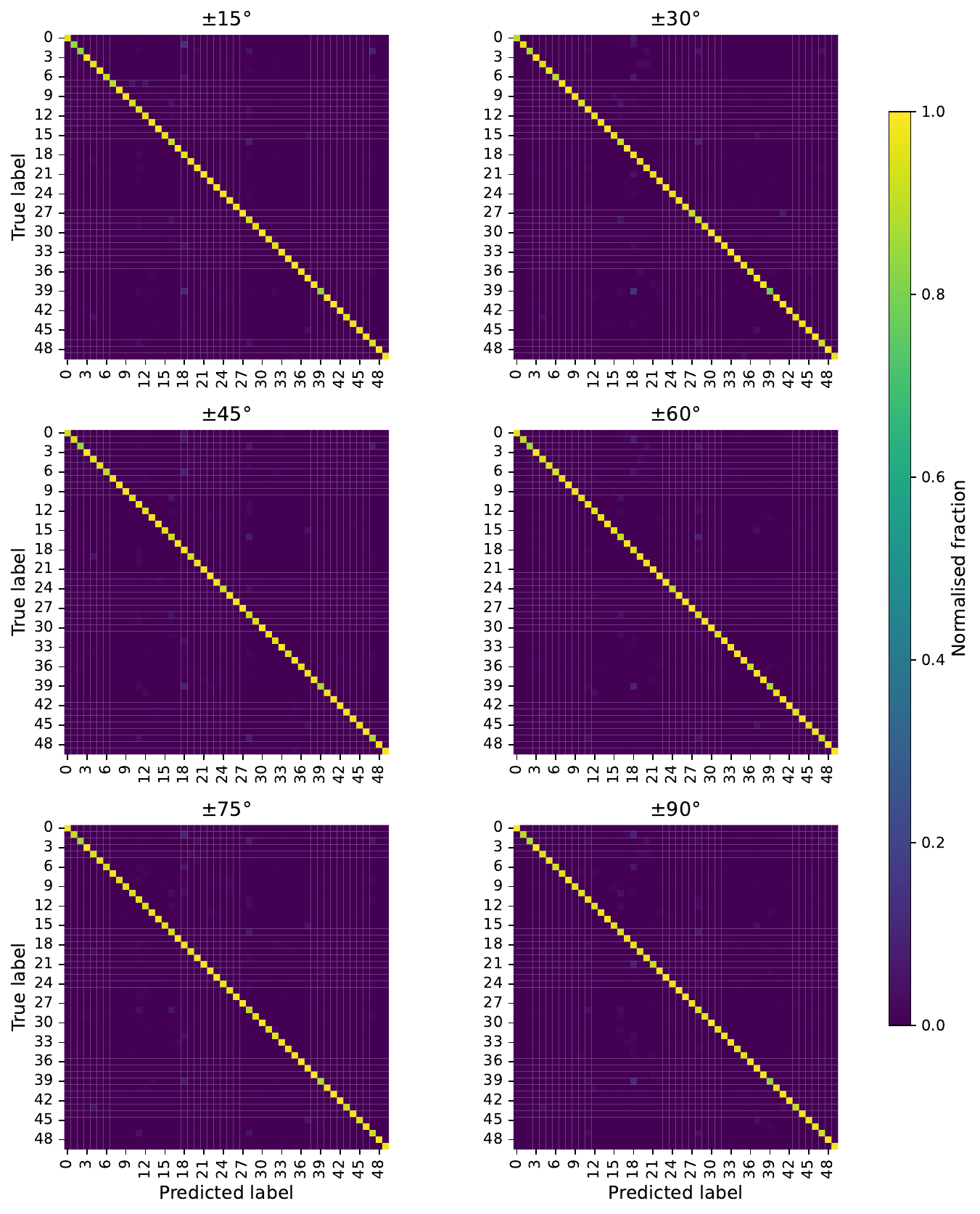}\\
    \caption{\textbf{Confusion matrices for each camera-angle bin in the CMU Multi-PIE dataset \cite{supp-4813399}, corresponding to the post-FRU classification results reported in Fig. 2.}}\label{S11}
  \end{center}
\end{figure}
\clearpage
\begin{figure}
  \begin{center}
  \includegraphics[width=\linewidth]{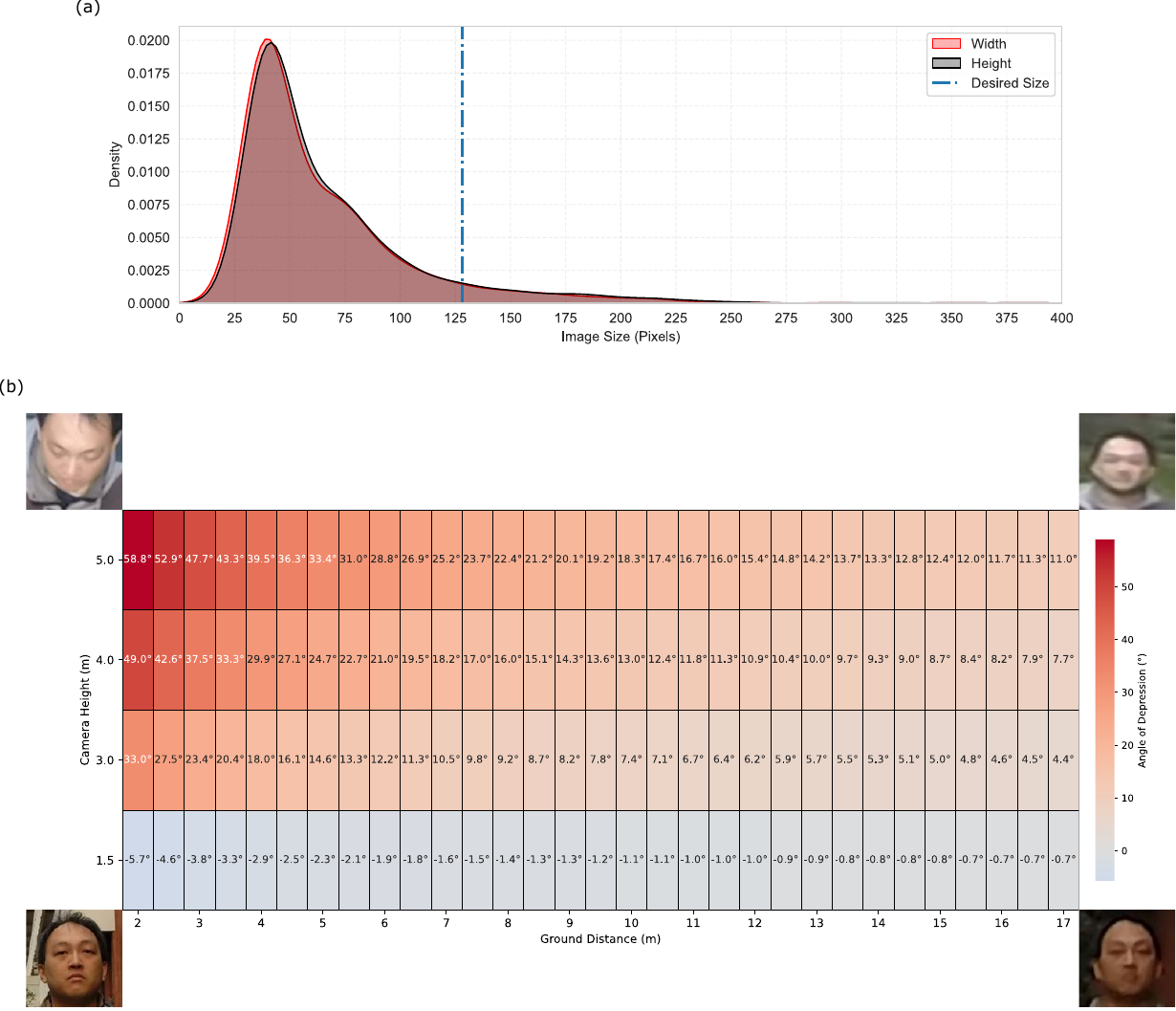}\\
    \caption{\textbf{Distributional characteristics of images in the DroneFace dataset \cite{supp-10.1145/3083187.3083214}.} \textbf{(a)} Kernel density estimate (KDE) plot of image dimensions in the original DroneFace dataset. The red curve represents the distribution of image widths, while the black curve represents image heights. A slight discrepancy between widths and heights indicates that not all images are square. It is evident that the majority of cropped images are well below $125\times125$ pixels. The blue vertical dashed line indicates the target resolution of 128 pixels. \textbf{(b)} Angle-of-depression table derived from average subject height and camera configuration. The angle varies with camera distance from the subject, ranging from 17\,m to 2\,m. Representative images are shown at the corners of the figure, corresponding to the nearest angle–distance settings in the table. The dataset reflects a real-world scenario with multiple degrees of freedom in the captured images.}
    \label{S12}
  \end{center}
\end{figure}
\clearpage
\begin{figure}
  \begin{center}
  \includegraphics[width=0.75\linewidth]{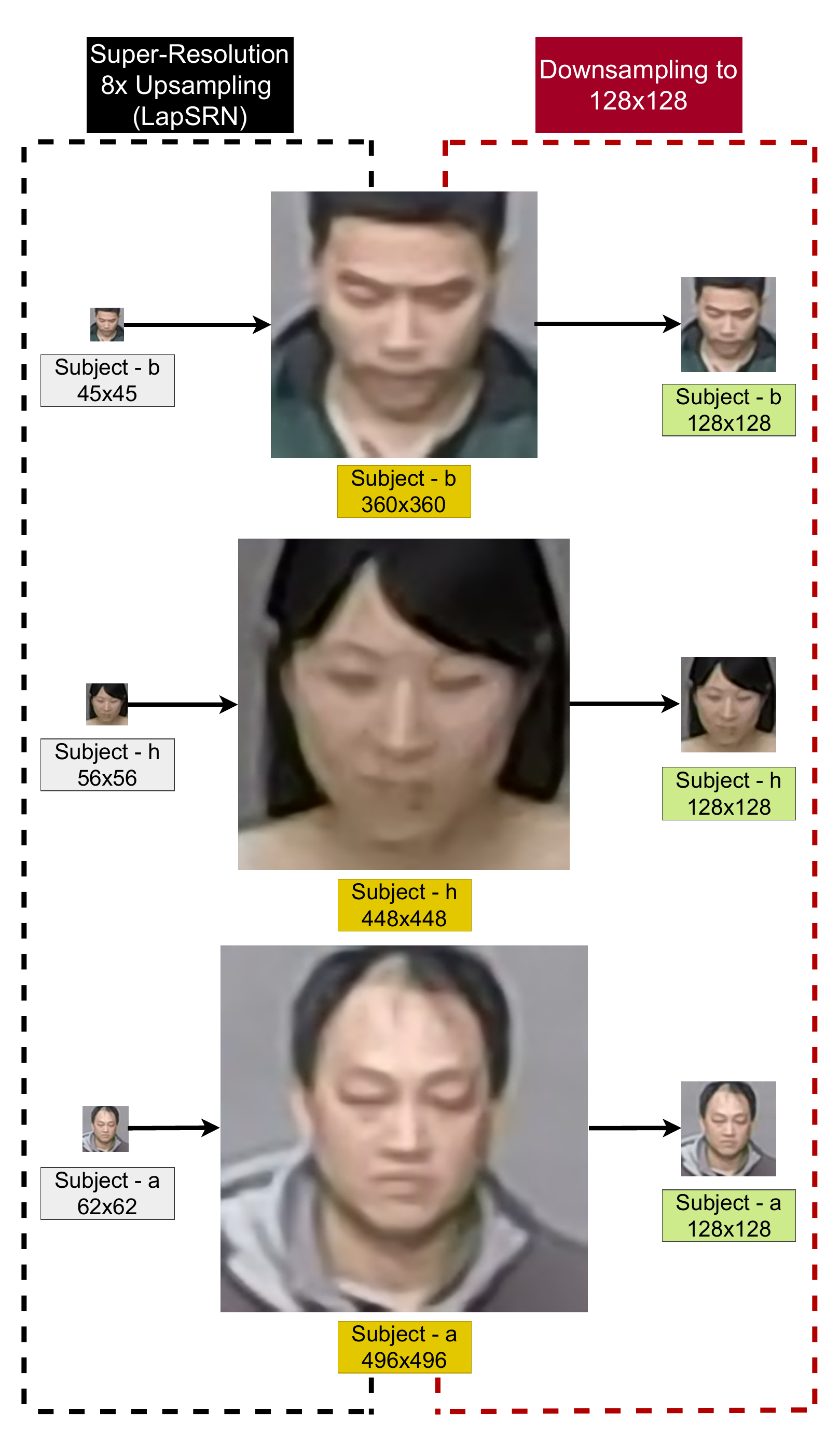}\\
  \caption{\textbf{Workflow for image resolution adjustment in the DroneFace dataset \cite{supp-10.1145/3083187.3083214} through upsampling and downsampling.} Small images on the left-hand side belong to the original DroneFace dataset. In the middle section of the figure, the super-resolution (SR) versions of \addtext{these} images are presented. These high-resolution (HR) images were processed using a deep learning-based SR model, LapSRN \cite{lai2017deeplaplacianpyramidnetworks}, significantly increasing their resolution by 8x, preserving content details and reducing pixelation. On the right-hand side, the HR images are then downsampled to a predetermined size of 128x128 pixels.}\label{S13}
  \end{center}
\end{figure}
\clearpage

\begin{figure}
  \begin{center}
  \includegraphics[width=\linewidth]{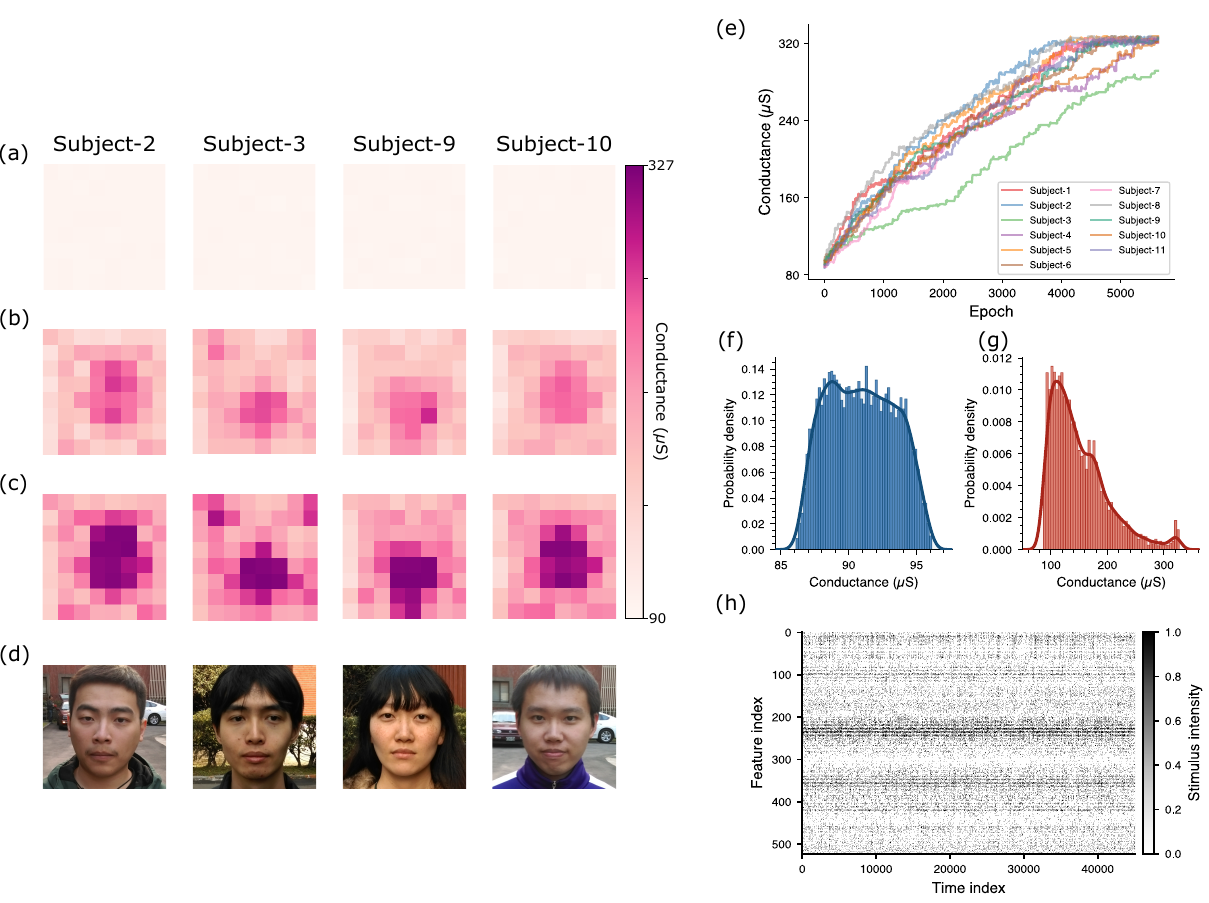}\\

  \caption{\textbf{Memristor-based classification dynamics on the DroneFace dataset \cite{supp-10.1145/3083187.3083214}.} \textbf{(a)} Memristor resistive states for the given subject classes before training. \textbf{(b)} Memristor resistive states at 50\% of training. \textbf{(c)} Memristor resistive states after training. \textbf{(d)} Representative ground-truth images of the corresponding subjects. \textbf{(e)} Learning curve of a synapse across all subject classes during training. \textbf{(f)} Initial conductance distribution before training. \textbf{(g)} Post-training conductance distribution. \textbf{(h)} Raster plot showing neuronal spike activity over time.}
\label{S14}
  \end{center}
\end{figure}

\clearpage

\begin{figure}
  \begin{center}
  \includegraphics[width=\linewidth]{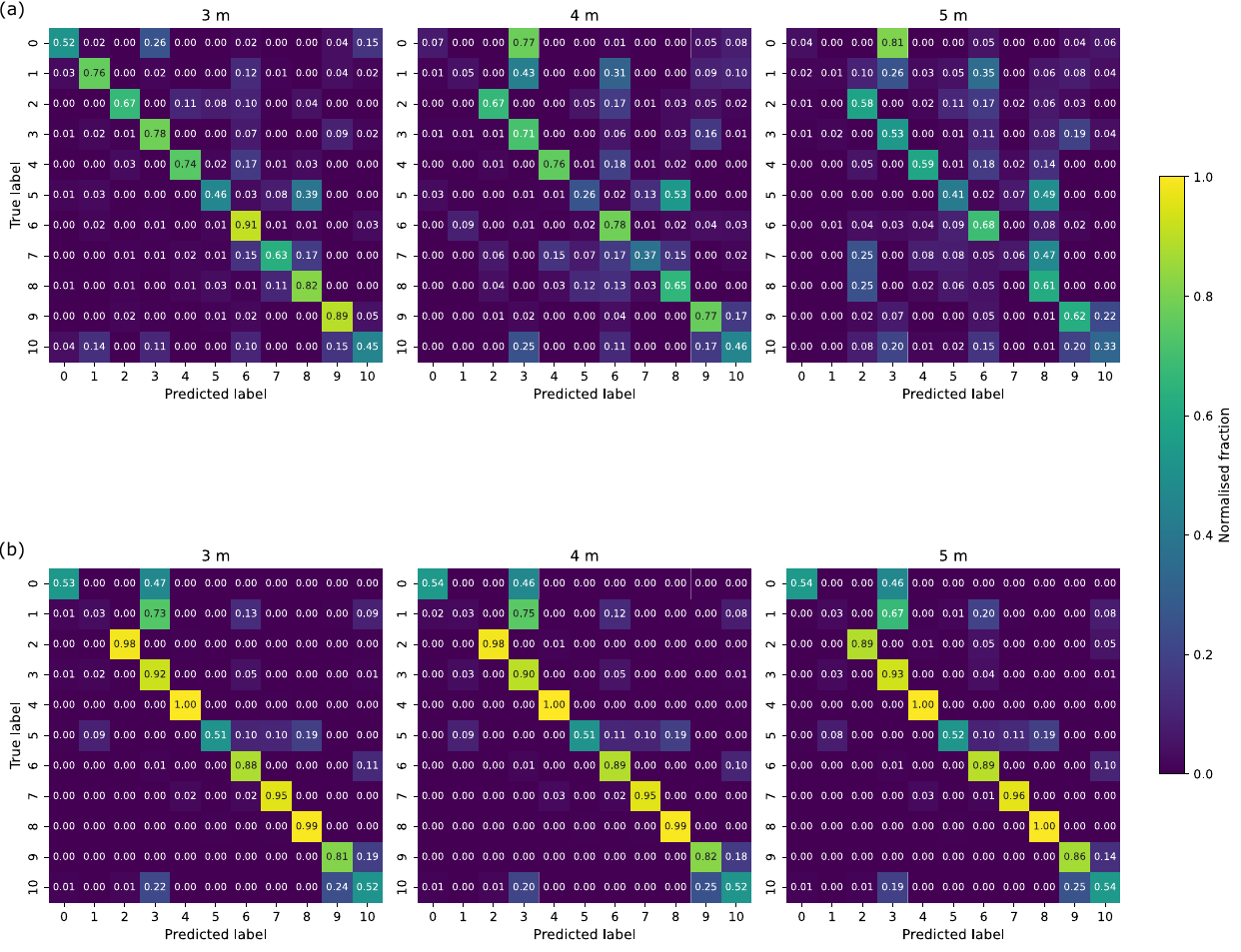}\\
    \caption{\textbf{Confusion matrices for each altitude setting of the DroneFace dataset \cite{supp-10.1145/3083187.3083214}, corresponding to the identification results shown in \addtext{Fig. 3}.} \textbf{(a)} Baseline. \textbf{(b)} After FRU reconstruction.}
    \label{S15}
  \end{center}
\end{figure}

\clearpage

\begin{figure}
  \begin{center}
  \includegraphics[width=0.999\linewidth]{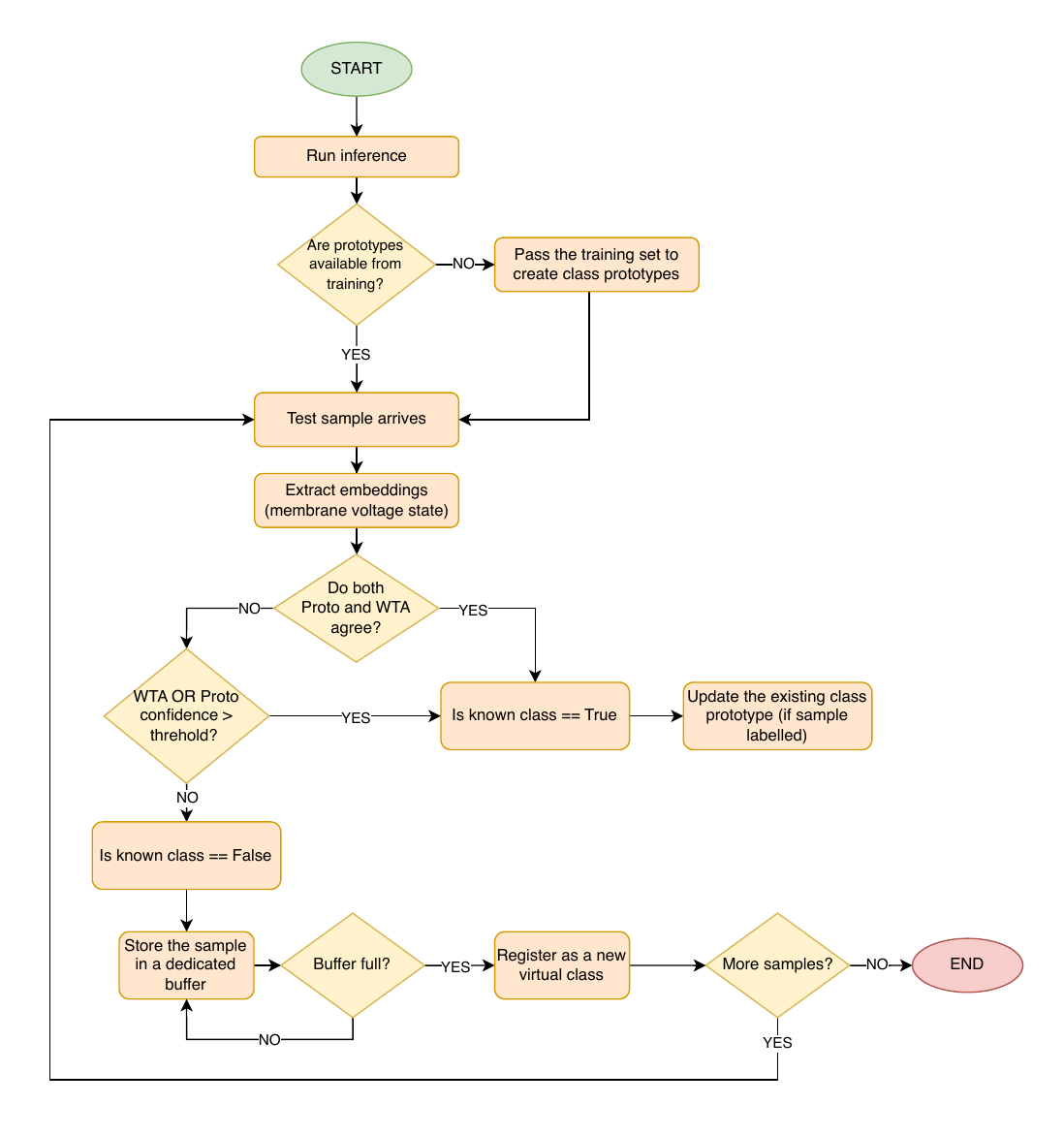}\\
    \caption{\textbf{Flowchart describing the prototype-based adaptive learning framework.}}
    \label{S16}
  \end{center}
\end{figure}

\clearpage

\begin{table}
\centering
\footnotesize
\renewcommand{\arraystretch}{1.4}
\setlength{\tabcolsep}{8pt}
\begin{tabular}{|c|c|}
\hline
\textbf{\small{Parameter}}                 & \textbf{\small{Value}}                                             \\ \hline
Random Seeds                       & 200-219 \\ \hline
\textbf{Features Extraction Core}           & \textbf{}                                         \\ \hline
In Features                         & 512                                                    \\ \hline
Out Features                         & 50                                                    \\ \hline
Number of Training Epochs                         & 2                                         \\ \hline
Training Batch Size                     & 16                                                   \\ \hline
Learning Rate                        & 0.001                                                   \\ \hline
\textbf{Features-to-Spikes Converter}           & \textbf{}                                         \\ \hline

Spike Encoding                     & Rate Coding                                                \\ \hline
Encoding Time Step                 & 8                                                         \\ \hline

\textbf{memristor crossbar FCL}           & \textbf{}                                         \\ \hline

Array Size                         & 160 x 160                                                    \\ \hline
Read Noise                         & 1\%                                                        \\ \hline

Neuron Model                       & LIF                                                        \\ \hline
Leakage                            & -0.3                                                       \\ \hline
Learning Rule                      & Backpropagation                                            \\ \hline
Learning Rate                      & 3.5 x $10^{-6}$                                             \\ \hline
Noise Scale                        & $10^{-6}$                                                        \\ \hline
Initial Resistance                         & $11$ k$\Omega$ $\pm500 \Omega$                                                        \\ \hline
\textbf{Memristor Model}           & \textbf{}                                                 \\ \hline
$A_p$, $A_n$, $t_p$, $t_n$, $a_{0p}$, $a_{0n}$, $a_{1p}$, $a_{1n}$ & 0.00392, -0.21832, 30799, 124076, 20571, 43099, 4.1838, 3.4624        \\ \hline
\textbf{Weight Updating}           & \textbf{}                                                 \\ \hline
Voltage (Positive)                          & 1.8, 2.0, 2.4, 2.0, 2.4, 2.8  \\ Voltage (Negative)    &  -2.7, -2.9, -3.0, -3.1, -3.2, -3.3                       \\
\hline
Pulse Width (Positive) & $3 \times 10^{-6}$, $5 \times 10^{-6}$, $10^{-5}$, $10^{-5}$, $2 \times 10^{-5}$, $3 \times 10^{-5}$ \\ Pulse Width (Negative) & $2 \times 10^{-6}$, $3 \times 10^{-6}$, $5 \times 10^{-6}$, $5 \times 10^{-6}$, $10^{-5}$, $10^{-5}$ \\ \hline
R-Tolerance                        & 0.1\%                                                      \\ \hline
Maximum Update Steps               & 8                                                          \\ \hline
\end{tabular}
\caption{\textbf{List of FCU parameters used in the experiments, with results shown in Fig. 1 and Fig. 2.}}
\label{table:Params1}
\end{table}

\vspace{-3em}

\begin{table}
\centering
\footnotesize
\renewcommand{\arraystretch}{1.4}
\setlength{\tabcolsep}{8pt}
\begin{tabular}{|c|c|}
\hline
\textbf{\small{Parameter}}                 & \textbf{\small{Value}}                                             \\ \hline
Random Seeds                       & 651 \\ \hline
Image Size                       & $128\times128$ \\ \hline
Batch Size                         & 64  \\ \hline
Number of Training Epochs      & 23                                         \\ \hline
Learning Rate (LR)                   & 0.00025 (Generator), 0.0001 (Discriminator)                                                   \\ \hline
Warm-up Epochs        & 5               \\ \hline
LR Decay Factor       & 0.02               \\ \hline
Number of Generator Features        & 32                                                   \\ \hline
Number of Discriminator Features        & 32                                                   \\ \hline
Adam $\beta_1, \beta_2$ & (0.5, 0.999)                                                   \\ \hline

\end{tabular}
\caption{\textbf{List of FRU parameters used in the experiments, with results shown in Fig. 2.}}
\label{table:Params2}
\end{table}

\begin{table}
\centering
\begin{tabular}{c c c c c c c c c c}
\hline
Phase & Epoch Range & Interp. Epochs & $\lambda_{1}$ & $\lambda_{2}$ & $\lambda_{3}$ & $\lambda_{4}$ & $\lambda_{5}$ & $\lambda_{6}$ & $\lambda_{7}$ \\
\hline
1 & 0--12 & -- & 0.001 & 1 & 0.3 & 0.1 & 0.0005 & 0.3 & 0.1\\
2 & 12+ & 6  & 0.003 & 0.6 & 0.05 & 0.05 & 0.0005 & 0.35 & 0.25
\end{tabular}
\caption{\textbf{Loss function weights for FRU training used in the experiments, with results shown in Fig. 2.}}
\label{table:Params3}
\end{table}

\clearpage

\begin{table}
\centering
\footnotesize
\renewcommand{\arraystretch}{1.4}
\setlength{\tabcolsep}{8pt}
\begin{tabular}{|c|c|}
\hline
\textbf{\small{Parameter}}                 & \textbf{\small{Value}}                                             \\ \hline
Random Seeds                       & 200-219 \\ \hline
\textbf{Features Extraction Core}           & \textbf{}                                         \\ \hline
In Features                         & 512                                                    \\ \hline
Out Features                         & 11                                                    \\ \hline
Number of Training Epochs                         & 8                                         \\ \hline
Training Batch Size                     & 16                                                   \\ \hline
Learning Rate                        & 0.001                                                   \\ \hline
\textbf{Features-to-Spikes Converter}           & \textbf{}                                         \\ \hline

Spike Encoding                     & Rate Coding                                                \\ \hline
Encoding Time Step                 & 8                                                         \\ \hline

\textbf{memristor crossbar FCL}           & \textbf{}                                         \\ \hline

Array Size                         & 76 x 76                                                    \\ \hline
Read Noise                         & 1\%                                                        \\ \hline

Neuron Model                       & LIF                                                        \\ \hline
Leakage                            & -0.3                                                       \\ \hline
Learning Rule                      & Backpropagation                                            \\ \hline
Learning Rate                      & 3.5 x $10^{-6}$                                             \\ \hline
Noise Scale                        & $10^{-6}$                                                        \\ \hline
Initial Resistance                         & $11$ k$\Omega$ $\pm500 \Omega$                                                        \\ \hline
\textbf{Memristor Model}           & \textbf{}                                                 \\ \hline
$A_p$, $A_n$, $t_p$, $t_n$, $a_{0p}$, $a_{0n}$, $a_{1p}$, $a_{1n}$ & 0.00392, -0.21832, 30799, 124076, 20571, 43099, 4.1838, 3.4624           \\ \hline
\textbf{Weight Updating}           & \textbf{}                                                 \\ \hline
Voltage (Positive)                          & 1.8, 2.0, 2.4, 2.0, 2.4, 2.8  \\ Voltage (Negative)    &  -2.7, -2.9, -3.0, -3.1, -3.2, -3.3                       \\
\hline
Pulse Width (Positive) & $3 \times 10^{-6}$, $5 \times 10^{-6}$, $10^{-5}$, $10^{-5}$, $2 \times 10^{-5}$, $3 \times 10^{-5}$ \\ Pulse Width (Negative) & $2 \times 10^{-6}$, $3 \times 10^{-6}$, $5 \times 10^{-6}$, $5 \times 10^{-6}$, $10^{-5}$, $10^{-5}$ \\ \hline
R-Tolerance                        & 0.1\%                                                      \\ \hline
Maximum Update Steps               & 8                                                          \\ \hline
\end{tabular}
\caption{\textbf{List of FCU parameters used in the experiments, with results shown in Fig. 3.}}
\label{table:Params4}
\end{table}

\vspace{-3em}

\begin{table}
\centering
\footnotesize 
\renewcommand{\arraystretch}{1.4} 
\setlength{\tabcolsep}{8pt}
\begin{tabular}{|c|c|}
\hline
\textbf{\small{Parameter}}                 & \textbf{\small{Value}}                                             \\ \hline
Random Seeds                       & 651 \\ \hline
Image Size                       & $128\times128$ \\ \hline
Batch Size                         & 64  \\ \hline
Number of Training Epochs      & 20                                         \\ \hline
Learning Rate (LR)                       & 0.0002 (Generator), 0.0001 (Discriminator)                                                   \\ \hline
Warm-up Epochs        & 4               \\ \hline
LR Decay Factor       & 0.04               \\ \hline
Number of Generator Features        & 16                                                   \\ \hline
Number of Discriminator Features        & 16                                                   \\ \hline
Adam $\beta_1, \beta_2$ & (0.5, 0.999)                                                   \\ \hline

\end{tabular}
\caption{\textbf{List of FRU parameters used in the experiments, with results shown in Fig. 3.}}
\label{table:Params5}
\end{table}

\begin{table}
\centering
\begin{tabular}{c c c c c c c c c c}
\hline
Phase & Epoch Range & Interp. Epochs & $\lambda_{1}$ & $\lambda_{2}$ & $\lambda_{3}$ & $\lambda_{4}$ & $\lambda_{5}$ & $\lambda_{6}$ & $\lambda_{7}$ \\
\hline
1 & 0--7 & -- & 0.01 & 1 & 0.2 & 0.1 & 0.0005 & 0.3 & 0.1\\
2 & 7--15 & 3  & 0.05 & 0.5 & 0.01 & 0.05 & 0.0005 & 0.3 & 0.3\\
3 & 15+    & 2  & 0.01 & 0.3 & 0.0 & 0.02 & 0.0005 & 0.25 & 0.25\\
\hline
\end{tabular}
\caption{\textbf{Loss function weights for FRU training used in the experiments, with results shown in Fig. 3.}}
\label{table:Params6}
\end{table}

\clearpage

\begin{table*}[t]
\centering
\begin{adjustbox}{width=1.15\textwidth,center}
\begin{threeparttable}
\scriptsize
\setlength{\tabcolsep}{3.5pt}
\renewcommand{\arraystretch}{1.1}
\begin{tabular}{llccccc>{\centering\arraybackslash}p{2.75cm}>{\centering\arraybackslash}p{2.75cm}}
\hline
\textbf{Coding} & \textbf{$T$} &
\makecell{\textbf{Accuracy}\\\textbf{(\%)} $\uparrow$} &
\makecell{\textbf{Inference}\\\textbf{Latency} $\downarrow$} &
\makecell{\textbf{Input Spikes}\\\textbf{/Img.} $\downarrow$} &
\makecell{\textbf{SOPs/Img.}\\\textbf{($\times10^3$)} $\downarrow$} &
\makecell{\textbf{Spike}\\\textbf{Rate (\%)} $\downarrow$} &
\makecell{\textbf{25\% Drop:}\\\textbf{Acc. / $\Delta$ (\%)} $\downarrow$} &
\makecell{\textbf{50\% Drop:}\\\textbf{Acc. / $\Delta$ (\%)} $\downarrow$} \\
\hline

\multirow{4}{*}{\raisebox{-7ex}{\makecell[c]{Burst}}}
& 2  & $61.8 \pm 27.2$ & $1.00 \pm 0.00$ & $87.0 \pm 19.0$   & $0.96 \pm 0.21$  & $7.9 \pm 1.7$  & $40.4 \pm 17.3$ / $(21.4 \pm 11.3)$ & $25.8 \pm 9.4$ / $(36.0 \pm 19.7)$ \\
& 4  & $55.9 \pm 13.2$ & $3.00 \pm 0.00$   & $213.3 \pm 36.0$  & $2.35 \pm 0.40$  & $9.7 \pm 1.6$  & $33.4 \pm 7.6$ / $(22.4 \pm 6.6)$  & $21.2 \pm 4.2$ / $(34.7 \pm 10.7)$ \\
& 8  & $42.5 \pm 12.4$ & $6.95 \pm 0.22$   & $448.8 \pm 70.4$  & $4.94 \pm 0.77$  & $10.2 \pm 1.6$ & $25.0 \pm 6.2$ / $(17.4 \pm 7.2)$  & $16.4 \pm 3.2$ / $(26.1 \pm 10.4)$ \\
& 16 & $29.6 \pm 10.3$ & $13.60 \pm 0.92$  & $910.0 \pm 139.3$ & $10.01 \pm 1.53$ & $10.3 \pm 1.5$ & $15.3 \pm 4.3$ / $(14.3 \pm 7.3)$  & $9.9 \pm 2.5$ / $(19.7 \pm 9.2)$ \\
\hline

\multirow{4}{*}{\raisebox{-7ex}{\makecell[c]{Phase}}}
& 2  & $80.5 \pm 35.7$ & $0.95 \pm 0.22$ & $87.0 \pm 19.0$    & $0.96 \pm 0.21$  & $7.9 \pm 1.7$  & $77.5 \pm 33.5$ / $(3.1 \pm 3.0)$  & $68.9 \pm 28.6$ / $(11.6 \pm 8.3)$ \\
& 4  & $98.1 \pm 1.3$  & $2.70 \pm 0.46$   & $316.4 \pm 39.7$   & $3.48 \pm 0.44$  & $14.3 \pm 1.8$ & $96.1 \pm 1.8$ / $(2.0 \pm 1.0)$   & $91.1 \pm 2.8$ / $(7.0 \pm 2.1)$ \\
& 8  & $97.3 \pm 1.7$  & $3.05 \pm 0.59$   & $899.2 \pm 77.2$   & $9.89 \pm 0.85$  & $20.2 \pm 1.7$ & $95.9 \pm 2.5$ / $(1.4 \pm 1.3)$   & $93.5 \pm 2.9$ / $(3.9 \pm 1.9)$ \\
& 16 & $97.4 \pm 2.3$  & $3.35 \pm 0.57$   & $2118.1 \pm 161.5$ & $23.30 \pm 1.78$ & $23.7 \pm 1.8$ & $96.6 \pm 2.4$ / $(0.8 \pm 0.9)$   & $94.8 \pm 2.8$ / $(2.5 \pm 1.4)$ \\
\hline

\multirow{4}{*}{\raisebox{-7ex}{\makecell[c]{Rate}}}
& 2  & $89.0 \pm 4.0$  & $1.00 \pm 0.00$ & $114.2 \pm 17.2$  & $1.26 \pm 0.19$  & $10.3 \pm 1.5$ & $83.2 \pm 4.6$ / $(5.8 \pm 1.4)$  & $72.6 \pm 5.8$ / $(16.4 \pm 3.2)$ \\
& 4  & $93.4 \pm 3.8$  & $1.40 \pm 0.49$ & $228.3 \pm 34.5$  & $2.51 \pm 0.38$  & $10.3 \pm 1.5$ & $90.5 \pm 3.9$ / $(2.9 \pm 1.2)$  & $83.0 \pm 5.4$ / $(10.5 \pm 3.1)$ \\
& \textbf{8}
& $\mathbf{95.2 \pm 2.7}$
& $\mathbf{2.25 \pm 0.62}$
& $\mathbf{457.0 \pm 68.9}$
& $\mathbf{5.03 \pm 0.76}$
& $\mathbf{10.3 \pm 1.5}$
& $\mathbf{93.3 \pm 3.0}$ / $\mathbf{(1.9 \pm 1.0)}$
& $\mathbf{88.6 \pm 4.0}$ / $\mathbf{(6.7 \pm 1.9)}$ \\
& 16 & $95.6 \pm 2.8$  & $2.45 \pm 0.92$ & $914.4 \pm 138.4$ & $10.06 \pm 1.52$ & $10.3 \pm 1.5$ & $94.3 \pm 3.6$ / $(1.3 \pm 1.1)$  & $90.6 \pm 4.4$ / $(5.0 \pm 2.6)$ \\
\hline

\end{tabular}
\begin{tablenotes}
\centering
\vspace{0.25em}
\item $\uparrow$ Higher values indicate better performance; $\downarrow$ lower values
indicate better performance.
\end{tablenotes}
\vspace{0.25em}
\caption{%
\leftskip=0pt
\rightskip=0pt
\parfillskip=0pt\addtext{\textbf{Coding-scheme comparison on FCU.} The model was trained on the same training set used for the experiments presented in Fig. 2, with images encoded using different coding schemes and numbers of steps \textbf{$T$} as listed above. For inference, the FRU-reconstructed images, originally captured at $\pm15^\circ$ from the CMU Multi-PIE dataset \cite{supp-4813399}, were encoded using the coding scheme and number of steps $T$ specified for each experiment in the table above. The table reports \textbf{Accuracy}, accuracy under \textbf{25\% and 50\% input-spike dropout}, \textbf{Inference Latency}, \addtext{\textbf{Input Spikes/Img.}}, \textbf{SOPs/Img.}, \addtext{and} \textbf{Spike Rate}. To ensure robustness and consistency, the benchmark was conducted using the same 20 random seeds as those listed in Supplementary Table \ref{table:Params1}. The configuration used in all recognition experiments is highlighted in bold.}}
    \vspace{2pt}
\begin{tablenotes}
\footnotesize
\item \addtext{\textit{Notes:} \textbf{Inference Latency} is the first time step (in units of $T$) at which the running classification decision reaches 95\% of its final accuracy, measuring how quickly the network commits to its final decision regardless of the final accuracy achieved. It is reported as a raw number of time steps rather than physical time. \textbf{SOPs/Img.} denotes network-wide synaptic operations per image, calculated as input spikes weighted by their fan-out. It is numerically identical to the input-layer-only synaptic operations because hidden and output neurons do not register a threshold-gated firing event for any tested $T$. \textbf{Spike Rate} denotes the average firing probability per neuron per time step, bounded in $[0,1]$ and therefore comparable across different $T$ values, unlike raw spike counts, which scale with $T$. At inference time, each input spike is independently dropped with probability $p \in \{0.25, 0.5\}$. \textbf{Accuracy (\%)} denotes the clean baseline accuracy (without dropout), while \textbf{Acc.~@ 25\%/50\% Drop} denotes the accuracy under the corresponding input-spike dropout level; the value in parentheses, $\Delta$, denotes the accuracy drop in percentage points relative to the clean baseline.}
\end{tablenotes}
\label{table:encoding_benchmark}

\end{threeparttable}
\end{adjustbox}
\end{table*}

\clearpage

\begin{table*}[t]
\centering
\begin{adjustbox}{width=1.2\textwidth,center}
\begin{threeparttable}
\small
\setlength{\tabcolsep}{8pt}
\renewcommand{\arraystretch}{2.5}
\begin{tabular}{
  >{\centering\arraybackslash}p{2.2cm}
  >{\centering\arraybackslash}p{2.7cm}
  >{\centering\arraybackslash}p{1.7cm}
  >{\centering\arraybackslash}p{3.7cm}
  >{\centering\arraybackslash}p{3.10cm}
  >{\centering\arraybackslash}p{2.05cm}
  >{\centering\arraybackslash}p{2.5cm}
  >{\centering\arraybackslash}p{2.9cm}
}
\toprule
\textbf{Work} &
\makecell{\textbf{Task}\\\textbf{Objective}} &
\textbf{Architecture} &
\textbf{Dataset} &
\makecell{\textbf{Input Pose}\\\textbf{Diversity}} &
\makecell{\textbf{Input Image}\\\textbf{Resolution}} &
\textbf{Parameters} &
\makecell{\textbf{Recognition}\\\textbf{Performance (\%)}} \\
\midrule
TP-GAN\cite{supp-8237529}        & Frontalisation                       & GAN  & Multi-PIE/LFW  & Yaw                                                                   & $128\times128\times3$ & 62.6M\tnote{a}                              & 98.68\tnote{b, c} \\
CAPG-GAN\cite{8578974}      & Frontalisation                       & GAN  & Multi-PIE/LFW  & Yaw                                                                   & $128\times128\times3$ & 67.9M\tnote{a}                              & 99.82\tnote{b, c} \\
DR-GAN\cite{Tran_2019}        & Frontalisation + Recognition          & GAN  & Multi-PIE      & Yaw                                                                   & $96\times96\times3$   & 58.3M\tnote{d}                              & 94.00\tnote{b, c} \\
\cmidrule(lr){2-7}
James et al.\cite{7865953}  & Recognition                  & HTM\tnote{e}  & Yale/AR/ORL & Frontal only\tnote{f} & Various\tnote{g}      & --                                 & 86.67\tnote{h} \\
Yao et al.\cite{Yao2017FaceCU}    & Recognition                           & SLP\tnote{i}  & Yale        & Frontal only                                                          & $20\times16\times1$   & $\sim$960                          & 91.70 \\
Wang et al.\cite{10.1021/acsami.8b16841}   & Recognition & MLP\tnote{j} & Yale & Frontal only & $32\times32\times1$ & $\sim$265.7k & 95.20 \\
\cmidrule(lr){2-7}
\textbf{This Work} & \textbf{Frontalisation + Recognition} & \textbf{GAN + SNN} & \textbf{Multi-PIE/DroneFace} & \textbf{Yaw/Pitch} & \textbf{$128\times128\times3$} & \textbf{10.9M (GAN) + 25.6k (SNN)} & 96.31\tnote{b} \\
\bottomrule
\end{tabular}

\addtext{\begin{tablenotes}
\scriptsize
\item [a] Values on the LFW dataset \cite{LIAO2025111112}.
\item[b] Rank-1 recognition rates on the Multi-PIE dataset for angle bin of ±15°.
\item[c] Adopted from Hu et al. \cite{8578974}.
\item[d] Adapted from Swetha et al. \cite{K2026}.
\item[e] Hierarchical Temporal Memory.
\item[f] Pose noted as "Frontal only" for AR and Yale subsets; ORL subset includes incidental, unquantified pose variation.
\item[g] Input resolution varies by dataset; not directly comparable across rows.
\item [h] Performance rate on the Yale dataset \cite{7865953}. \
\item[i] Sparse/Single-Layer Perceptron.
\item[j] Multi-Layer Perceptron.
\end{tablenotes}}
\vspace{0.5em}

\caption{%
\leftskip=0pt
\rightskip=0pt
\parfillskip=0pt\normalsize\addtext{\textbf{Overview of related face frontalisation/recognition works.} The table presents architectures designed for different purposes. The first three works employ Generative Adversarial Networks (GANs) for frontal face synthesis, while the subsequent three works adopt various lightweight algorithms on memristor-based architectures for the face recognition task. This work combines both directions, using a GAN for frontalisation and a Spiking Neural Network (SNN) for recognition.}}
\label{table:related-work}

\end{threeparttable}
\end{adjustbox}
\end{table*}

\clearpage

\section*{Supplementary Note 1: Dataset preprocessing}

The proposed application was evaluated on two different datasets: CMU Multi-PIE \cite{supp-4813399} and DroneFace \cite{supp-10.1145/3083187.3083214}. The Multi-PIE dataset comprises images captured under a tightly controlled setup, resulting in high consistency across different yaw angles. In contrast, the DroneFace dataset reflects a more practical, less controlled environment. This distinction is further illustrated in Figure \ref{S12}(b), which presents the distribution of angle of depression, defined as the angle between the horizontal line from the drone and the line of sight down to the subject on the ground, for each camera height and ground distance pair, provided in the supplementary material. Moreover, the DroneFace dataset is significantly smaller, containing only 1,364 images across 11 subjects, whereas Multi-PIE includes more than 750,000 images of 337 subjects. Given the DroneFace dataset’s limited size, variability, and lower consistency, preprocessing and data augmentation are essential to enhance data robustness and diversity, enabling effective use and reliable evaluation of the model.

The raw images in the DroneFace were methodically acquired from a range of specific distance and height combinations. For each subject, the drone captured images at a certain altitude and distances starting from 17 metres to 2 metres from the subject, with increments of 0.5 metres between each step. The same approach was repeated for different height settings (1.5, 3, 4, and 5 metres above ground). This applies to each subject across the dataset. Subsequently, 1,364 facial images were extracted from 620 raw images. Depending on the distance-altitude setting of each raw image, the size of the cropped images varies. Closer distances to subjects yield larger cropped images with higher resolutions. Figure \ref{S12}(a) demonstrates how image sizes are distributed, providing insights into the diversity and variability of dimensions within the original dataset. According to the figure, only a very small amount of images, 98 out of 1,364 ($\approx 7.2\%$), are larger than $128\times128$ pixels, which is the desired size for the application. 

To address the dataset's limited size and variability in spatial resolution, super-resolution and data augmentation techniques were applied to the original images. Using the smallest image ($23\times31$ pixels) as a reference, all smaller images were upsampled 8× with the pre-trained \textit{LapSRN} \cite{lai2017deeplaplacianpyramidnetworks} model to ensure consistency. The resulting super-resolution images were then downsampled to $128\times128$ pixels via bilinear interpolation. Figure \ref{S13} in the supplementary material provides a detailed illustration of the super-resolution process.

After correcting the spatial dimensions of the images, a series of data augmentation techniques was applied, including slight adjustments to brightness, contrast, the introduction of Gaussian noise, and geometric transformations such as rotation, shifting, and scaling. Consequently, a more robust and comprehensive dataset was obtained for use in the different stages of the proposed facial classification framework.
\clearpage

\section*{\addtext{Supplementary Note 2: Generator loss functions}}
\label{supp:gen_loss}

The generator is trained using a multi-objective loss that combines reconstruction fidelity, perceptual quality, identity preservation, and structural coherence. These objectives jointly encourage the synthesis of high-quality frontal face images while maintaining identity consistency. 

The adversarial training is formulated as a minimax game between the generator and the discriminator. The discriminator is trained as a binary classifier that distinguishes real frontal images \(x\) from generated samples \(\hat{x} = G(I^S)\). The discriminator loss is defined as:

\begin{equation}
\mathcal{L}_D =
\frac{1}{2} \left[
\mathcal{L}_{\text{BCE}}(D(x), y_{\text{gt}})
+
\mathcal{L}_{\text{BCE}}(D(\hat{x}), y_{\text{gen}})
\right],
\end{equation}

where \(y_{\text{gt}} = 0.9\) and \(y_{\text{gen}} = 0.0\) denote label-smoothed targets for ground-truth and generated frontal images, respectively, which improve training stability and mitigate overconfidence in the discriminator. The term \(\mathcal{L}_{\text{BCE}}\) denotes the binary cross-entropy loss applied on logits for numerical stability.

The generator is trained to fool the discriminator by encouraging generated samples to be classified as real. Its adversarial objective is given by:
\begin{equation}
\mathcal{L}_G^{\text{adv}} =
\mathcal{L}_{\text{BCE}}(D(\hat{x}), 1.0).
\end{equation}

This formulation encourages the generator to produce realistic frontal images that are indistinguishable from ground-truth samples, while the use of label smoothing stabilises adversarial training and mitigates overconfident discriminator outputs.

To ensure spatial fidelity between generated and ground-truth images, two complementary pixel-level losses are employed. The L1 loss (also referred to as Mean Absolute Error) is defined as:

\begin{equation}
\mathcal{L}_{L1} = \mathbb{E}\left[ \| G(I^S) - x \|_1 \right]
\end{equation}

where \(G(I^S)\) denotes the generated frontal image and \(x\) denotes the ground-truth frontal image. This loss is particularly effective at preventing blurring \addtext{artefacts} by encouraging pixel-wise accuracy. Complementarily, the L2 loss (also referred to as Mean Squared Error) is defined as:

\begin{equation}
\mathcal{L}_{L2} = \mathbb{E}\left[ \| G(I^S) - x \|_2^2 \right]
\end{equation}

L2 loss emphasises larger errors, providing additional regularisation and ensuring that significant discrepancies between the generated and ground-truth images are penalised more heavily.

To ensure structural coherence in the generated frontal images, a symmetry loss is introduced. This loss encourages the generator to produce faces with symmetric left and right halves, a natural property of frontal-facing faces. The symmetry loss is defined as:

\begin{equation}
\mathcal{L}_{\text{sym}} = \mathbb{E}\left[
1 - \cos\left(
\mathrm{vec}(G(I^S)_{L}),
\mathrm{vec}(\mathrm{flip}(G(I^S)_{R}))
\right)
\right]
\end{equation}

where \(G(I^S)_{L}\) and \(G(I^S)_{R}\) denote the left and right halves of the generated frontal image, respectively. \(\mathrm{flip}(\cdot)\) denotes horizontal flipping, and \(\mathrm{vec}(\cdot)\) vectorises the spatial and channel dimensions into a single vector. This soft regularisation preserves natural facial asymmetries (e.g., expressions, eye shape) while discouraging strong structural inconsistencies. Additionally, a total variation loss is applied to reduce high-frequency noise and promote smooth transitions in the generated image. The total variation loss is defined as:

\begin{equation}
\mathcal{L}_{\text{tv}} = \frac{1}{2} \left(
\left\| \nabla_y \hat{I} \right\|_1 +
\left\| \nabla_x \hat{I} \right\|_1
\right),
\quad \text{where } \hat{I} = G(I^S).
\end{equation}

where $\nabla_x$ and $\nabla_y$ denote the horizontal and vertical finite-difference gradient operators, respectively, and $\|\cdot\|_1$ denotes the element-wise L1 norm computed over all spatial locations of the image. This loss penalises local intensity variations, encouraging spatial smoothness in the generated output.

While the loss functions defined thus far form the core of the FRU's behaviour, additional auxiliary losses are introduced to enhance the framework's generalisation capability. These auxiliary losses improve the model's perceptual understanding, though they come with increased computational cost during training. Specifically, identity preservation and perceptual losses are incorporated to provide crucial inductive biases, enabling the model to better capture semantic content and maintain identity consistency across variations.

A primary goal in face reconstruction is to maintain identity consistency between the input and output images, which is accomplished by using an identity loss based on ArcFace \cite{Deng_2019_CVPR} embeddings, and it is computed as:

\begin{equation}
    \mathcal{L}_{\text{id}} = 1 - E_{\text{gen}} \cdot E_{\text{gt}}
\end{equation}

where \(E_{\text{gen}}\) and \(E_{\text{gt}}\) are L2-normalised ArcFace embeddings of the generated and ground-truth images, respectively. This loss measures cosine distance between L2-normalised ArcFace embeddings and is computed after resizing both images to the ArcFace input size (112×112). While identity loss ensures semantic consistency by preserving identity features, perceptual loss focuses on maintaining broader semantic content beyond identity alone:

\begin{equation}
\mathcal{L}_{\text{perc}} = \sum_{l \in \mathcal{L}} \|F_l(G(I^S)) - F_l(x)\|_1
\end{equation}

where \(F_l\) denotes the feature extraction function at layer \(l\) of a pre-trained VGG-16 \cite{simonyan2015deepconvolutionalnetworkslargescale} network, and \(\mathcal{L}\) includes the layers \texttt{relu2\_2} and \texttt{relu3\_3}. This encourages the generator to produce images that are perceptually similar to the ground-truth in terms of high-level semantic content and texture, rather than relying solely on pixel-wise similarity.

The generator's total loss is a weighted combination of all components:

\begin{equation}
    \mathcal{L}_G = \lambda_{1} \mathcal{L}_G^{\text{adv}} + \lambda_{2} \mathcal{L}_{L1} + \lambda_{3} \mathcal{L}_{L2} + \lambda_{4} \mathcal{L}_{\text{sym}} + \lambda_{5} \mathcal{L}_{\text{tv}} +\lambda_{6} \mathcal{L}_{\text{id}} + \lambda_{7} \mathcal{L}_{\text{perc}}
\end{equation}

To \addtext{coordinate the} competing objectives across different stages of training, a three-phase loss weighting schedule is adopted\addtext{, motivated by a coarse-to-fine optimisation strategy. Reconstruction losses are emphasised during the initial stage to establish the global facial structure, while perceptual and identity constraints are progressively strengthened during later stages to refine high-level visual quality and identity consistency. The weighting coefficients are empirically selected following common practice in the literature \cite{supp-8237529, 8578974}.} The specific weighting coefficients used in experiments are reported in Supplementary Tables \ref{table:Params3} and \ref{table:Params6}.

\clearpage

\section*{\addtext{Supplementary Note 3: Memristor switching rate model and fitting variables}}

As described in previous work \cite{messaris2017compactverilogareramswitching}, the general state equation of a memristor is linked to the applied bias voltage $v$ and the resistance state $R$, and is formulated as:
\begin{equation}
\frac{dR}{dt} = s(v)\, f(R, v)
\end{equation}

where $s(v)$ is the switching sensitivity function and $f(R, v)$ is the window function, which together govern the evolution of the device's internal state variable.

The switching sensitivity function $s(v)$ is defined as:
\begin{equation}
s(v) =
\begin{cases}    
A_{p}\left(-1 + e^{\frac{|v|}{t_{p}}}\right), & v > 0 \\
A_{n}\left(-1 + e^{\frac{|v|}{t_{n}}}\right), & v < 0
\end{cases}
\end{equation}

where $A_{p}$ and $A_{n}$ are the switching scaling factors for the positive and negative voltage regimes, and $t_{p}$ and $t_{n}$ are the corresponding voltage threshold parameters.

The window function $f(R, v)$ is defined as:
\begin{equation}
f(R, v) =
\begin{cases}
\left(a_{0,p} + a_{1,p}v - R\right)^2, & v > 0 \\
\left(R - a_{0,n} + a_{1,n}v\right)^2, & v < 0
\end{cases}
\end{equation}

where $a_{0,p,n}$ and $a_{1,p,n}$ are empirical fitting parameters that capture asymmetry and non-linearity in the voltage-dependent resistance evolution under positive and negative bias conditions.

Rearranging the above expressions, the voltage-dependent switching rate can be written as:
\begin{equation}
\frac{dR}{dt} = m(R, v) =
\begin{cases}
A_{p}\left(-1 + e^{\frac{|v|}{t_{p}}}\right)\left(a_{0,p} + a_{1,p}v - R\right)^2, & v > 0 \\
A_{n}\left(-1 + e^{\frac{|v|}{t_{n}}}\right)\left(R - a_{0,n} + a_{1,n}v\right)^2, & v < 0
\end{cases}
\end{equation}

\clearpage

\section*{\addtext{Supplementary Note 4: Face classification unit algorithms}}

\vspace{-2em}

\begin{algorithm}[H]
\caption{Training path}
\label{alg:hardware_training}
\begin{algorithmic}[1]
\footnotesize
 
\Require Training spike sequences $\mathcal{X}$, labels $\mathcal{Y}$, number of training samples $E$ ($=|\mathcal{X}|$),
number of simulation steps per image $T$, learning rate $\eta$, 
membrane leakage $\lambda$, firing threshold $\theta$,
resistance tolerance $\epsilon_R$, weight scale $\alpha$,
device resistance bounds $R_{\min}, R_{\max}$
(conductance bounds $G_{\min}=R_{\max}^{-1}$, $G_{\max}=R_{\min}^{-1}$),
normalisation switch $\beta \in \{0,1\}$,
candidate pulse sets $\mathcal{P}_{\mathrm{pos}}, \mathcal{P}_{\mathrm{neg}}$, output layer $O$, hidden layer $H$
\Ensure Programmed synaptic weights $W$, frozen test weights $W_{\mathrm{test}}$
 
\Statex \textbf{Definitions}
\Statex \hspace{\algorithmicindent}
$\operatorname{Normalize}_\beta(G)=\begin{cases}\operatorname{clip}\!\left(\alpha\,\dfrac{G-G_{\min}}{G_{\max}-G_{\min}},\ 0,\ 1\right) & \beta=1\\[4pt] G & \beta=0\end{cases}$
\Statex \hspace{\algorithmicindent}
$\operatorname{DeNormalize}_\beta(p)=\begin{cases}\left(p\,\dfrac{G_{\max}-G_{\min}}{\alpha}+G_{\min}\right)^{-1} & \beta=1\\[4pt] p^{-1} & \beta=0\end{cases}$
\Statex \hspace{\algorithmicindent}
\Comment{$\beta$ is the config flag \texttt{NORMALISE}}
 
\Statex \addtext{\textbf{Initialisation}}
 
\For{each device-mapped synapse $(i,j)$}
    \State $R_{ij} \gets \operatorname{ReadDevice}(i,j)$
    \State $W_{ij} \gets \operatorname{Normalize}_\beta(R_{ij}^{-1})$
\EndFor
 
\Statex \textbf{Training}
 
\For{each training sample $e=1,\ldots,E$}
    \State Reset membrane potentials: $V_j \gets 0$ for all hidden and output neurons $j$
 
    \For{$t=1,\ldots,T$}
 
        \State $V_j(t) \gets (1-\lambda)\,V_j(t-1) + \sum_i \sigma_{ij}\, S_i(t)\, W_{ij}$
 
        \Statex \hspace{\algorithmicindent}\hspace{\algorithmicindent}
        where $\sigma_{ij}=+1$ (excitatory) or $-1$ (inhibitory); neurons are
        visited in index order, so an earlier neuron's $S_i(t)$ can drive a
        later neuron's $V_j(t)$ within this same $t$
 
        \State $S_j(t) \gets \mathbb{I}\!\left[V_j(t)>\theta\right]$
        \State $S_j^{\mathrm{free}}(t) \gets S_j(t)$, \quad $S_O^{\mathrm{free}}(t) \gets \{S_j^{\mathrm{free}}(t)\}_{j \in O}$
        \Statex \hspace{\algorithmicindent}\hspace{\algorithmicindent}
        \Comment{fire state before winner-take-all (WTA) arbitration}
 
        \If{$\sum_{j \in O} S_j^{\mathrm{free}}(t) > 1$}
            \State $j^{*} \gets \arg\max_{j \in O} V_j(t)$
            \State $S_O(t) \gets \operatorname{onehot}(j^{*})$
        \Else
            \State $S_O(t) \gets S_O^{\mathrm{free}}(t)$
        \EndIf
 
        \State Store the resulting spike state $S(t)$
 
    \EndFor
 
    \State $\hat{y}_e \gets S_O(T)$
    \Statex \hspace{\algorithmicindent}\Comment{output spike pattern at the final time step}
    \State Record training prediction and accuracy
 
    \If{$e = E$}
        \State \textbf{continue} \Comment{last sample: no backward pass}
    \EndIf
 
    \State $\displaystyle \hat{p}_j \gets \frac{\exp\!\big(V_j(T)\,S_j(T)\big)}{\sum_{k\, \in\, O}\exp\!\big(V_k(T)\,S_k(T)\big)}$ \quad for $j \in O$
    \Statex \hspace{\algorithmicindent}\Comment{softmax over the output layer}
 
    \For{each neuron $j$ in reverse network order}
        \Statex \hspace{\algorithmicindent}\Comment{output layer first (highest index), then hidden layer}
 
        \If{$j \in O$}
            \If{$S_j(T) = y_j$}
                \State \textbf{continue}
            \EndIf
            \If{$|V_j(T)-\theta| > \theta$}
                \State $\delta_j \gets (\hat{p}_j - y_j)\,S_j^{\mathrm{free}}(T)$
            \Else
                \State $\delta_j \gets (\hat{p}_j - y_j)\,\dfrac{S_j^{\mathrm{free}}(T) + \tfrac{1}{2}V_j(T)}{\theta}$
            \EndIf
 
        \ElsIf{$j \in H$}
            \State $\phi'(V_j) \gets \dfrac{1}{2\theta}$ if $|V_j(T)-\theta|\le\theta$, else $0$
            \State $\displaystyle \delta_j \gets \phi'(V_j)\sum_{k\, \in\, \mathrm{post}(j)} \delta_k\, W_{jk}^{(e)}$
            \Statex \hspace{\algorithmicindent}\hspace{\algorithmicindent}
 
        \Else
            \State \textbf{continue} \Comment{input neuron - nothing to backpropagate}
        \EndIf
 
        \If{$\delta_j \ne 0$}
 
            \For{each presynaptic connection $(i,j)$}
 
                \State $\Delta W_{ij} \gets -\eta\,\delta_j\,S_i^{\mathrm{free}}(T)$
                \Statex \hspace{\algorithmicindent}\hspace{\algorithmicindent}\hspace{\algorithmicindent}
                \Comment{pre-WTA free-fire state at the image's final time step}
                \State $p_{\mathrm{target}} \gets \operatorname{Normalize}_\beta(R_{ij}^{-1}) + \Delta W_{ij}$
                \State $R_{\mathrm{target}} \gets \operatorname{DeNormalize}_\beta(p_{\mathrm{target}})$
                \State $R_{ij} \gets \Call{SelectAndProgram}{R_{ij}, R_{\mathrm{target}}, \epsilon_R}$
                \Statex \hspace{\algorithmicindent}\hspace{\algorithmicindent}\hspace{\algorithmicindent}
                \Comment{Algorithm~\ref{alg:select_program}}
                \State $W_{ij} \gets \operatorname{Normalize}_\beta(R_{ij}^{-1})$
 
            \EndFor
 
        \EndIf
 
    \EndFor
 
\EndFor
 
\State Freeze final measured weights: $W_{\mathrm{test}} \gets W$
 
\end{algorithmic}
\end{algorithm}

\begin{algorithm}[t]
\caption{SelectAndProgram - closed-loop pulse programming (subroutine of
Algorithm~\ref{alg:hardware_training})}
\label{alg:select_program}
\begin{algorithmic}[1]
\footnotesize

\Require Current resistance $R$, target resistance $R_{\mathrm{target}}$,
tolerance $\epsilon_R$, candidate pulse sets
$\mathcal{P}_{\mathrm{pos}}=\{(V_k,t_{pw,k})\}$,
$\mathcal{P}_{\mathrm{neg}}=\{(V_k,t_{pw,k})\}$,
integration step $\Delta t$, maximum steps $\mathrm{maxSteps}$,
device parameters $A_p,A_n,a_{0p},a_{1p},a_{0n},a_{1n},t_p,t_n$
\Ensure Updated resistance $R$ after closed-loop programming,
with the tolerance satisfied or $\mathrm{maxSteps}$ reached

\Statex \textbf{Device pulse-response model} ($H(\cdot)$: Heaviside step)
\Statex \hspace{\algorithmicindent}
$r(V)=
\begin{cases}
a_{0p}+a_{1p}V & V>0\\
a_{0n}+a_{1n}V & V\leq 0
\end{cases}$
\qquad
$s(V)=
\begin{cases}
A_p\!\left(e^{|V|/t_p}-1\right) & V>0\\
A_n\!\left(e^{|V|/t_n}-1\right) & V\leq 0
\end{cases}$

\Statex \hspace{\algorithmicindent}
$f(R,V)=
\begin{cases}
H\!\big(r(V)-R\big)\big(r(V)-R\big)^2 & V>0\\
H\!\big(R-r(V)\big)\big(R-r(V)\big)^2 & V\leq 0
\end{cases}$

\Function{PredictResistance}{$R,V,t_{pw}$}
    \For{$k=1,\ldots,\lfloor t_{pw}/\Delta t\rfloor$}
        \State $R \gets R+s(V)\,f(R,V)\,\Delta t$
    \EndFor
    \State \Return $R$
\EndFunction

\Function{SelectAndProgram}{$R,R_{\mathrm{target}},\epsilon_R$}
    \State $\mathrm{steps}\gets0$
    \While{$\dfrac{|R-R_{\mathrm{target}}|}{R_{\mathrm{target}}}
    >\epsilon_R$ \textbf{and} $\mathrm{steps}<\mathrm{maxSteps}$}

        \State $\mathcal{P}\gets\mathcal{P}_{\mathrm{neg}}$ if
        $R>R_{\mathrm{target}}$ (resistance must decrease), else
        $\mathcal{P}_{\mathrm{pos}}$

        \State $\displaystyle
        (V^{*},t_{pw}^{*})\gets
        \arg\min_{(V,t_{pw})\in\mathcal{P}}
        \left|\Call{PredictResistance}{R,V,t_{pw}}
        -R_{\mathrm{target}}\right|$
        \Comment{Select the closest predicted state}

        \State $\operatorname{ProgramDevice}(V^{*},t_{pw}^{*})$
        \Comment{Apply the selected pulse}

        \State $R\gets\operatorname{ReadDevice}()$
        \State $\mathrm{steps}\gets\mathrm{steps}+1$

    \EndWhile
    \State \Return $R$
\EndFunction

\end{algorithmic}
\end{algorithm}

\begin{algorithm}[t]
\caption{Test-time inference}
\label{alg:inference}
\begin{algorithmic}[1]
\footnotesize

\Require Test spike sequences $\mathcal{X}_{\mathrm{test}}$,
frozen weights $W_{\mathrm{test}}$,
prototype memory $\mathcal{P}_{\mathrm{proto}}$,
distance threshold $\tau_{\mathrm{proto}}$

\Ensure Predicted class labels and classification metrics

\For{each test sample $e=1,\ldots,E_{\mathrm{test}}$}

    \State Reset membrane potentials: $V_j \gets 0$

    \State $\hat{y}_e \gets
    \operatorname{ForwardPass}(\mathcal{X}_{\mathrm{test},e}, W_{\mathrm{test}})$
    \Statex \hspace{\algorithmicindent}
    \Comment{As in Algorithm~\ref{alg:hardware_training}'s per-time-step
    block: leaky integration, thresholding, winner-take-all (WTA);
    no backward pass and no device writes}

    \State Record test prediction and accuracy

    \If{prototype memory is enabled}
        \Statex \hspace{\algorithmicindent}
        \Comment{Flag a sample as a possible unseen class when its
        embedding lies far from every known-class prototype, rather than
        always trusting the WTA vote. The full rule also considers WTA
        confidence and newly registered classes.}

        \State $d \gets
        \operatorname{dist}\!\left(
        \mathrm{embedding}_e,\,
        \text{nearest prototype in }\mathcal{P}_{\mathrm{proto}}
        \right)$

        \If{$d > \tau_{\mathrm{proto}}$}
            \State Apply prototype-based class decision
        \EndIf

        \If{online adaptation is enabled \textbf{and} ground-truth label
        $y_e$ is available}
            \Statex \hspace{\algorithmicindent}
            \Comment{Update the prototype using an EMA. The update
            strength can be confidence-dependent, e.g., strong when both
            WTA and prototype predictions agree with $y_e$, and moderate
            when only one agrees.}

            \State $\mathrm{proto}_{y_e} \gets
            \operatorname{normalize}\!\left(
            \alpha\,\mathrm{embedding}_e
            + (1-\alpha)\,\mathrm{proto}_{y_e}
            \right)$
        \EndIf

    \EndIf

\EndFor

\State Compute test accuracy
\State Save prototype-memory snapshot

\end{algorithmic}
\end{algorithm}

\clearpage

\clearpage

\section*{Supplementary Note 5: Adaptive learning protocol}

Image embeddings were derived from the membrane potential state of output neurons, normalised by the spike train duration. For each training image, the membrane potential accumulation of the output layer was extracted at the end of the spike train interval and normalised by the spike train duration, yielding a \addtext{biologically inspired}, time-normalised embedding representation.

The membrane potential of neuron $j$ evolves according to a leaky integrate-and-fire (LIF) dynamics model:

\begin{equation}
\mathcal{N}_j(t) =
\begin{cases}
0, & t \equiv 0 \pmod{T},\\[6pt]
(1-\lambda)\,\mathcal{N}_j(t-1)
+\displaystyle\sum_{i \in \mathrm{pre}(j)} s_i(t)\, w_{ij},
& \text{otherwise}.
\end{cases}
\end{equation}

where $\mathcal{N}_j(t)$ is the membrane potential of neuron $j$ at timestep $t$, \addtext{$\lambda$} is the leakage factor, $s_i(t) \in \{0,1\}$ is the spike from presynaptic neuron $i$, $\mathrm{pre}(j)$ is the set of presynaptic neurons connected to neuron $j$, $w_{ij}$ is the synaptic weight, and $T$ is the spike train duration. The membrane potential is reset to zero at image boundaries to ensure embeddings capture activity during the current image presentation only.

At the end of each image presentation, the embedding vector is defined as:
\begin{equation}
\mathbf{e}_n = \frac{\boldsymbol{\mathcal{N}}_{\mathrm{out}}(t^*)}{T}, 
\quad t^* = (n+1)T - 1,
\end{equation}
where $\boldsymbol{\mathcal{N}}_{\mathrm{out}}(t^*)$ denotes the output-layer membrane potential vector at time $t^*$, $T$ is the spike train duration, and $\mathbf{e}_n$ is the resulting time-normalised embedding.

Embeddings corresponding to a given class are aggregated to form the initial class prototype, computed as the mean of the class-specific embeddings. Let $\{\mathbf{e}_i^{(k)}\}_{i=1}^{M_k}$ denote the embeddings of training samples belonging to class $k$, where $\mathbf{e}_i^{(k)} \in \mathbb{R}^{d}$ is the normalised embedding of the $i$-th sample from class $k$ and $d$ denotes the dimensionality of the output layer. The initial class prototype is defined as:
\begin{equation}
\mathbf{p}_k =
\frac{1}{M_k}
\sum_{i=1}^{M_k} \mathbf{e}_i^{(k)} ,
\label{eq:prototype_def}
\end{equation}
where $M_k$ denotes the number of training samples in class $k$.

Image embeddings are extracted in the same manner during both training and inference, as defined above. The only distinction lies in the synaptic weights: during training, embeddings are computed using evolving weights that are updated throughout the learning process, whereas during inference, embeddings are computed using the fixed weights obtained at the final training epoch.

During inference, the model updates the prototypes of existing classes using labelled incoming test samples, enabling continual adaptation via confidence-gated prototype updates. The confidence-gating mechanism modulates the update strength according to the agreement between a prototype-based classifier and the winner-take-all (WTA) decision: when both mechanisms select the same class and the predictive confidence is high, prototype updates are applied with larger strength; when they disagree or confidence is low, updates are weakened or suppressed. This selective gating stabilises learned prototypes while allowing refinement from high-confidence samples. \addtext{A comparable principle is observed in memristive circuits realising classical conditioning, where associative strength is revised through graded, history-dependent modulation, supporting the biological plausibility of this gated update strategy \cite{11389851}.}

Prototypes are updated during online learning via exponential moving average (EMA):
\begin{equation}
\mathbf{p}_k \leftarrow
\frac{
\alpha\, \mathbf{e}_{test} + (1-\alpha)\, \mathbf{p}_k
}{
\left\|
\alpha\, \mathbf{e}_{test} + (1-\alpha)\, \mathbf{p}_k
\right\|_2 + \epsilon
}
\end{equation}

where $\alpha \in [0,1]$ is the EMA smoothing factor, $\mathbf{e}_{test}$ is the test embedding, and $\epsilon = 10^{-10}$ ensures numerical stability. The parameter $\alpha$ is adaptively modulated based on prediction confidence to prevent destabilisation while preserving previously learned knowledge.

The prototype-based classification approach employs cosine distance for decision-making, defined as:

\begin{equation}
d_{\mathrm{cos}}(\mathbf{e}_1, \mathbf{e}_2) =
1 - \frac{\mathbf{e}_1^\top \mathbf{e}_2}{\lVert \mathbf{e}_1 \rVert_2 \, \lVert \mathbf{e}_2 \rVert_2}
\end{equation}

where $\mathbf{e}_1, \mathbf{e}_2 \in \mathbb{R}^n$ denote embedding vectors. The cosine distance quantifies the angular separation between vectors in the embedding space, taking values in the interval $[0, 2]$, where $0$ indicates identical orientation and $2$ indicates diametrically opposed directions. This metric is invariant to vector magnitude, which is desirable when embeddings represent normalised firing rates.

An embedding is classified as unknown when its distance to the nearest known prototype exceeds a detection threshold $\delta$. Specifically, an embedding $\mathbf{e}_n$ is marked as unknown if
\begin{equation}
d_{\min}(\mathbf{e}_n) = \min_{k=1}^{K} d_{cos}(\mathbf{e}_n, \mathbf{p}_k) > \delta,
\end{equation}
where $K$ is the number of initially known classes. Upon detection, the system assigns the embedding to a virtual buffer indexed by $v \geq K$ and initiates a buffering process to accumulate evidence for potential class registration.

\addtext{The buffering mechanism groups unknown embeddings based on similarity, under the assumption that instances of the same novel class produce embeddings within a bounded region of the embedding space. The admission radius $\varepsilon$ is calibrated empirically from the network's own training data, as the $p$-th percentile of pairwise cosine distances between embeddings of the same known class, pooled across all $K$ known classes:
}

\begin{equation}
\varepsilon = \mathrm{percentile}_{p}\Big(\big\{\, d_{\mathrm{cos}}\big(\mathbf{e}_i^{(k)}, \mathbf{e}_j^{(k)}\big) \;:\; k = 1,\dots,K,\; i<j \,\big\}\Big),
\end{equation}

\addtext{computed once after training completes. This grounds the density radius in the network's own empirical intra-identity spread. Before calibration has run (e.g. transiently at cold start), a fallback radius of $1.5\,\delta$ is used in its place.
}

\addtext{When a new unknown embedding $\mathbf{e}_m$ is encountered, the system searches for an existing buffer $\mathcal{B}_v$ containing at least one previously buffered embedding within distance $\varepsilon$:}

\begin{equation}
\min_{i} d_{\mathrm{cos}}\big(\mathbf{e}_m, \mathbf{e}_i^{(v)}\big) \le \varepsilon .
\end{equation}

\addtext{If such a buffer exists, $\mathbf{e}_m$ is added to it; otherwise, a new buffer is created for a candidate novel class. This admission test is deliberately permissive: the more rigorous density check is applied once, at registration time, over the whole assembled buffer, described next.
}

\addtext{Registration of a virtual buffer as a permanent class requires that it first accumulate at least $n_{\mathrm{buffer}}$ embeddings, at which point density-based clustering (DBSCAN \cite{1996kddm.conf..226E}) is applied over the buffer's contents, using the same radius $\varepsilon$. A buffered embedding $\mathbf{e}_i^{(v)}$ is a \emph{core point} if at least $n_{\mathrm{minPts}}$ other embeddings in the buffer lie within $\varepsilon$ of it:
}

\begin{equation}
\big|\{\, j \neq i : d_{\mathrm{cos}}(\mathbf{e}_i^{(v)}, \mathbf{e}_j^{(v)}) \le \varepsilon \,\}\big| \;\ge\; n_{\mathrm{minPts}} .
\end{equation}

\addtext{Here $n_{\mathrm{minPts}}$ counts neighbouring points excluding the point itself. Core points are linked into clusters by density-reachability (mutual core-point adjacency within $\varepsilon$). Each remaining non-core point that lies within $\varepsilon$ of one or more core points is assigned to the cluster of its nearest core point; points with no core point within $\varepsilon$ are marked as noise. Let $\mathcal{C}_1, \mathcal{C}_2, \dots$ denote the resulting clusters (each comprising both core points and any border points attached to it), ordered by size, and let $n_v$ be the total number of embeddings in the buffer. The largest cluster $\mathcal{C}_1$ is registered as a new permanent class only if it accounts for at least a fraction $\rho_{\min}$ of the buffer, requiring that a substantial majority of the buffer be absorbed into this one coherent cluster before it is trusted as evidence for a new identity:
}
\begin{equation}
\frac{|\mathcal{C}_1|}{n_v} \;\ge\; \rho_{\min},
\label{eq:threshold}
\end{equation}

\addtext{with the class prototype initialised from the mean of $\mathcal{C}_1$'s embeddings (Eq. \ref{eq:prototype_def}) and all embeddings not in $\mathcal{C}_1$, including noise points and any smaller clusters, excluded from that prototype. If a second cluster $\mathcal{C}_2$ is also found within the same buffer, it is registered separately as an independent new class rather than merged into $\mathcal{C}_1$; this directly guards against a single buffer having accumulated a mixture of two distinct novel identities, since such a mixture is now split rather than averaged into one incoherent prototype. If no cluster in the buffer satisfies Eq. \ref{eq:threshold}, the buffer is retained unregistered and continues to accumulate future unknown embeddings that fall within it.
}

\clearpage

\putbib[supp_refs]

\end{bibunit}

%% file: sn-article.bbl

\begin{thebibliography}{16}
\ifx \bisbn   \undefined \def \bisbn  #1{ISBN #1}\fi
\ifx \binits  \undefined \def \binits#1{#1}\fi
\ifx \bauthor  \undefined \def \bauthor#1{#1}\fi
\ifx \batitle  \undefined \def \batitle#1{#1}\fi
\ifx \bjtitle  \undefined \def \bjtitle#1{#1}\fi
\ifx \bvolume  \undefined \def \bvolume#1{\textbf{#1}}\fi
\ifx \byear  \undefined \def \byear#1{#1}\fi
\ifx \bissue  \undefined \def \bissue#1{#1}\fi
\ifx \bfpage  \undefined \def \bfpage#1{#1}\fi
\ifx \blpage  \undefined \def \blpage #1{#1}\fi
\ifx \burl  \undefined \def \burl#1{\textsf{#1}}\fi
\ifx \doiurl  \undefined \def \doiurl#1{\url{https://doi.org/#1}}\fi
\ifx \betal  \undefined \def \betal{\textit{et al.}}\fi
\ifx \binstitute  \undefined \def \binstitute#1{#1}\fi
\ifx \binstitutionaled  \undefined \def \binstitutionaled#1{#1}\fi
\ifx \bctitle  \undefined \def \bctitle#1{#1}\fi
\ifx \beditor  \undefined \def \beditor#1{#1}\fi
\ifx \bpublisher  \undefined \def \bpublisher#1{#1}\fi
\ifx \bbtitle  \undefined \def \bbtitle#1{#1}\fi
\ifx \bedition  \undefined \def \bedition#1{#1}\fi
\ifx \bseriesno  \undefined \def \bseriesno#1{#1}\fi
\ifx \blocation  \undefined \def \blocation#1{#1}\fi
\ifx \bsertitle  \undefined \def \bsertitle#1{#1}\fi
\ifx \bsnm \undefined \def \bsnm#1{#1}\fi
\ifx \bsuffix \undefined \def \bsuffix#1{#1}\fi
\ifx \bparticle \undefined \def \bparticle#1{#1}\fi
\ifx \barticle \undefined \def \barticle#1{#1}\fi
\bibcommenthead
\ifx \bconfdate \undefined \def \bconfdate #1{#1}\fi
\ifx \botherref \undefined \def \botherref #1{#1}\fi
\ifx \url \undefined \def \url#1{\textsf{#1}}\fi
\ifx \bchapter \undefined \def \bchapter#1{#1}\fi
\ifx \bbook \undefined \def \bbook#1{#1}\fi
\ifx \bcomment \undefined \def \bcomment#1{#1}\fi
\ifx \oauthor \undefined \def \oauthor#1{#1}\fi
\ifx \citeauthoryear \undefined \def \citeauthoryear#1{#1}\fi
\ifx \endbibitem  \undefined \def \endbibitem {}\fi
\ifx \bconflocation  \undefined \def \bconflocation#1{#1}\fi
\ifx \arxivurl  \undefined \def \arxivurl#1{\textsf{#1}}\fi
\csname PreBibitemsHook\endcsname

\bibitem[\protect\citeauthoryear{Gross et~al.}{2008}]{supp-4813399}
\begin{bchapter}
\bauthor{\bsnm{Gross}, \binits{R.}},
\bauthor{\bsnm{Matthews}, \binits{I.}},
\bauthor{\bsnm{Cohn}, \binits{J.}},
\bauthor{\bsnm{Kanade}, \binits{T.}},
\bauthor{\bsnm{Baker}, \binits{S.}}:
\bctitle{Multi-pie}.
In: \bbtitle{2008 8th IEEE International Conference on Automatic Face \& Gesture Recognition},
pp. \bfpage{1}--\blpage{8}
(\byear{2008}).
\doiurl{10.1109/AFGR.2008.4813399}
\end{bchapter}
\endbibitem

\bibitem[\protect\citeauthoryear{Hsu and Chen}{2017}]{supp-10.1145/3083187.3083214}
\begin{bchapter}
\bauthor{\bsnm{Hsu}, \binits{H.-J.}},
\bauthor{\bsnm{Chen}, \binits{K.-T.}}:
\bctitle{Droneface: An open dataset for drone research}.
In: \bbtitle{Proceedings of the 8th ACM on Multimedia Systems Conference}.
\bsertitle{MMSys'17},
pp. \bfpage{187}--\blpage{192}.
\bpublisher{Association for Computing Machinery},
\blocation{New York, NY, USA}
(\byear{2017}).
\doiurl{10.1145/3083187.3083214} .
\burl{https://doi.org/10.1145/3083187.3083214}
\end{bchapter}
\endbibitem

\bibitem[\protect\citeauthoryear{Lai et~al.}{2017}]{lai2017deeplaplacianpyramidnetworks}
\begin{botherref}
\oauthor{\bsnm{Lai}, \binits{W.-S.}},
\oauthor{\bsnm{Huang}, \binits{J.-B.}},
\oauthor{\bsnm{Ahuja}, \binits{N.}},
\oauthor{\bsnm{Yang}, \binits{M.-H.}}:
Deep Laplacian Pyramid Networks for Fast and Accurate Super-Resolution
(2017).
\url{https://arxiv.org/abs/1704.03915}
\end{botherref}
\endbibitem

\bibitem[\protect\citeauthoryear{Huang et~al.}{2017}]{supp-8237529}
\begin{bchapter}
\bauthor{\bsnm{Huang}, \binits{R.}},
\bauthor{\bsnm{Zhang}, \binits{S.}},
\bauthor{\bsnm{Li}, \binits{T.}},
\bauthor{\bsnm{He}, \binits{R.}}:
\bctitle{Beyond face rotation: Global and local perception gan for photorealistic and identity preserving frontal view synthesis}.
In: \bbtitle{2017 IEEE International Conference on Computer Vision (ICCV)},
pp. \bfpage{2458}--\blpage{2467}
(\byear{2017}).
\doiurl{10.1109/ICCV.2017.267}
\end{bchapter}
\endbibitem

\bibitem[\protect\citeauthoryear{Hu et~al.}{2018}]{8578974}
\begin{bchapter}
\bauthor{\bsnm{Hu}, \binits{Y.}},
\bauthor{\bsnm{Wu}, \binits{X.}},
\bauthor{\bsnm{Yu}, \binits{B.}},
\bauthor{\bsnm{He}, \binits{R.}},
\bauthor{\bsnm{Sun}, \binits{Z.}}:
\bctitle{Pose-guided photorealistic face rotation}.
In: \bbtitle{2018 IEEE/CVF Conference on Computer Vision and Pattern Recognition},
pp. \bfpage{8398}--\blpage{8406}
(\byear{2018}).
\doiurl{10.1109/CVPR.2018.00876}
\end{bchapter}
\endbibitem

\bibitem[\protect\citeauthoryear{Tran et~al.}{2019}]{Tran_2019}
\begin{barticle}
\bauthor{\bsnm{Tran}, \binits{L.}},
\bauthor{\bsnm{Yin}, \binits{X.}},
\bauthor{\bsnm{Liu}, \binits{X.}}:
\batitle{Representation learning by rotating your faces}.
\bjtitle{IEEE Transactions on Pattern Analysis and Machine Intelligence}
\bvolume{41}(\bissue{12}),
\bfpage{3007}--\blpage{3021}
(\byear{2019})
\doiurl{10.1109/tpami.2018.2868350}
\end{barticle}
\endbibitem

\bibitem[\protect\citeauthoryear{James et~al.}{2017}]{7865953}
\begin{barticle}
\bauthor{\bsnm{James}, \binits{A.P.}},
\bauthor{\bsnm{Fedorova}, \binits{I.}},
\bauthor{\bsnm{Ibrayev}, \binits{T.}},
\bauthor{\bsnm{Kudithipudi}, \binits{D.}}:
\batitle{Htm spatial pooler with memristor crossbar circuits for sparse biometric recognition}.
\bjtitle{IEEE Transactions on Biomedical Circuits and Systems}
\bvolume{11}(\bissue{3}),
\bfpage{640}--\blpage{651}
(\byear{2017})
\doiurl{10.1109/TBCAS.2016.2641983}
\end{barticle}
\endbibitem

\bibitem[\protect\citeauthoryear{Yao et~al.}{2017}]{Yao2017FaceCU}
\begin{botherref}
\oauthor{\bsnm{Yao}, \binits{P.}},
\oauthor{\bsnm{Wu}, \binits{H.}},
\oauthor{\bsnm{Gao}, \binits{B.}},
\oauthor{\bsnm{Eryilmaz}, \binits{S.B.}},
\oauthor{\bsnm{Huang}, \binits{X.}},
\oauthor{\bsnm{Zhang}, \binits{W.}},
\oauthor{\bsnm{Zhang}, \binits{Q.}},
\oauthor{\bsnm{Deng}, \binits{N.}},
\oauthor{\bsnm{Shi}, \binits{L.}},
\oauthor{\bsnm{Wong}, \binits{H.-S.P.}},
\oauthor{\bsnm{Qian}, \binits{H.}}:
Face classification using electronic synapses.
Nature Communications
\textbf{8}
(2017)
\end{botherref}
\endbibitem

\bibitem[\protect\citeauthoryear{Wang et~al.}{2018}]{10.1021/acsami.8b16841}
\begin{barticle}
\bauthor{\bsnm{Wang}, \binits{T.-Y.}},
\bauthor{\bsnm{He}, \binits{Z.-Y.}},
\bauthor{\bsnm{Liu}, \binits{H.}},
\bauthor{\bsnm{Chen}, \binits{L.}},
\bauthor{\bsnm{Zhu}, \binits{H.}},
\bauthor{\bsnm{Sun}, \binits{Q.-Q.}},
\bauthor{\bsnm{Ding}, \binits{S.-J.}},
\bauthor{\bsnm{Zhou}, \binits{P.}},
\bauthor{\bsnm{Zhang}, \binits{D.W.}}:
\batitle{Flexible electronic synapses for face recognition application with multimodulated conductance states}.
\bjtitle{ACS Applied Materials \& Interfaces}
\bvolume{10}(\bissue{43}),
\bfpage{37345}--\blpage{37352}
(\byear{2018})
\doiurl{10.1021/acsami.8b16841}
{\href{https://arxiv.org/abs/https://pubs.acs.org/aamick/article-pdf/10/43/37345/8380702/am8b16841.pdf}{{https://pubs.acs.org/aamick/article-pdf/10/43/37345/8380702/am8b16841.pdf}}}
\end{barticle}
\endbibitem

\bibitem[\protect\citeauthoryear{Liao et~al.}{2025}]{LIAO2025111112}
\begin{barticle}
\bauthor{\bsnm{Liao}, \binits{J.}},
\bauthor{\bsnm{Guha}, \binits{T.}},
\bauthor{\bsnm{Sanchez}, \binits{V.}}:
\batitle{Self-supervised random mask attention gan in tackling pose-invariant face recognition}.
\bjtitle{Pattern Recognition}
\bvolume{159},
\bfpage{111112}
(\byear{2025})
\doiurl{10.1016/j.patcog.2024.111112}
\end{barticle}
\endbibitem

\bibitem[\protect\citeauthoryear{K et~al.}{2026}]{K2026}
\begin{botherref}
\oauthor{\bsnm{K}, \binits{S.}},
\oauthor{\bsnm{Darmanayagam}, \binits{S.E.}},
\oauthor{\bsnm{Cyril}, \binits{S.R.R.}}:
Mmf-cyclegan: A multi-scale generative framework for robust and identity-preserving face frontalization.
CMES - Computer Modeling in Engineering and Sciences
\textbf{146}(3)
(2026)
\doiurl{10.32604/cmes.2026.077293}
\end{botherref}
\endbibitem

\bibitem[\protect\citeauthoryear{Deng et~al.}{2019}]{Deng_2019_CVPR}
\begin{bchapter}
\bauthor{\bsnm{Deng}, \binits{J.}},
\bauthor{\bsnm{Guo}, \binits{J.}},
\bauthor{\bsnm{Xue}, \binits{N.}},
\bauthor{\bsnm{Zafeiriou}, \binits{S.}}:
\bctitle{Arcface: Additive angular margin loss for deep face recognition}.
In: \bbtitle{Proceedings of the IEEE/CVF Conference on Computer Vision and Pattern Recognition (CVPR)}
(\byear{2019})
\end{bchapter}
\endbibitem

\bibitem[\protect\citeauthoryear{Simonyan and Zisserman}{2015}]{simonyan2015deepconvolutionalnetworkslargescale}
\begin{botherref}
\oauthor{\bsnm{Simonyan}, \binits{K.}},
\oauthor{\bsnm{Zisserman}, \binits{A.}}:
Very Deep Convolutional Networks for Large-Scale Image Recognition
(2015).
\url{https://arxiv.org/abs/1409.1556}
\end{botherref}
\endbibitem

\bibitem[\protect\citeauthoryear{Messaris et~al.}{2017}]{messaris2017compactverilogareramswitching}
\begin{botherref}
\oauthor{\bsnm{Messaris}, \binits{I.}},
\oauthor{\bsnm{Serb}, \binits{A.}},
\oauthor{\bsnm{Khiat}, \binits{A.}},
\oauthor{\bsnm{Nikolaidis}, \binits{S.}},
\oauthor{\bsnm{Prodromakis}, \binits{T.}}:
A compact Verilog-A ReRAM switching model
(2017).
\url{https://arxiv.org/abs/1703.01167}
\end{botherref}
\endbibitem

\bibitem[\protect\citeauthoryear{Gao et~al.}{2026}]{11389851}
\begin{barticle}
\bauthor{\bsnm{Gao}, \binits{S.}},
\bauthor{\bsnm{Shi}, \binits{F.}},
\bauthor{\bsnm{Xu}, \binits{X.}},
\bauthor{\bsnm{Chen}, \binits{J.}},
\bauthor{\bsnm{Iu}, \binits{H.H.-C.}},
\bauthor{\bsnm{Cao}, \binits{Y.}},
\bauthor{\bsnm{Qin}, \binits{L.}},
\bauthor{\bsnm{Mou}, \binits{J.}}:
\batitle{A memristor-based neural network circuit with classical conditioning and fear generalization}.
\bjtitle{IEEE Transactions on Consumer Electronics}
\bvolume{72}(\bissue{2}),
\bfpage{3213}--\blpage{3224}
(\byear{2026})
\doiurl{10.1109/TCE.2026.3663359}
\end{barticle}
\endbibitem

\bibitem[\protect\citeauthoryear{{Ester} et~al.}{1996}]{1996kddm.conf..226E}
\begin{bchapter}
\bauthor{\bsnm{{Ester}}, \binits{M.}},
\bauthor{\bsnm{{Kriegel}}, \binits{H.-P.}},
\bauthor{\bsnm{{Sander}}, \binits{J.}},
\bauthor{\bsnm{{Xu}}, \binits{X.}}:
\bctitle{{A Density-Based Algorithm for Discovering Clusters in Large Spatial Databases with Noise}}.
In: \beditor{\bsnm{{Pfitzner}}, \binits{D.W.}},
\beditor{\bsnm{{Salmon}}, \binits{J.K.}} (eds.)
\bbtitle{Second International Conference on Knowledge Discovery and Data Mining (KDD'96). Proceedings of a Conference Held August 2-4},
pp. \bfpage{226}--\blpage{331}
(\byear{1996})
\end{bchapter}
\endbibitem

\end{thebibliography}
